\documentclass{article}

\usepackage[preprint]{neurips_2026}

\usepackage[utf8]{inputenc} % allow utf-8 input
\usepackage[T1]{fontenc}    % use 8-bit T1 fonts
\usepackage{hyperref}       % hyperlinks
\usepackage{url}            % simple URL typesetting
\usepackage{booktabs}       % professional-quality tables
\usepackage{amsfonts}       % blackboard math symbols
\usepackage{nicefrac}       % compact symbols for 1/2, etc.
\usepackage{microtype}      % microtypography
\usepackage{xcolor}         % colors
\usepackage{graphicx}
\usepackage{amsmath}

\usepackage[capitalize,noabbrev]{cleveref}
\usepackage{subcaption}

\usepackage{color,soul}
\usepackage{xcolor}
\definecolor{cerulean}{rgb}{0.0,0.48,0.65}
\definecolor{green}{rgb}{0.01, 0.75, 0.24}
\definecolor{Black}{RGB}{0.0, 0.0, 0.0}

\newcommand{\shadow}[1]{}
\def\s{\shadow}

\title{TabPFN-MT: A Natively Multitask In-Context Learner for Tabular Data}

\author{%
  Cormac Cureton \\
  McGill University\\
  Mila - Quebec AI Institute \\
  Montreal, QC, Canada \\
  \texttt{cormac.cureton@mail.mcgill.ca} \\
  \And
  Narges Armanfard \\
  McGill University\\
  Mila - Quebec AI Institute \\
  Montreal, QC, Canada \\
  \texttt{narges.armanfard@mcgill.ca} \\
}

\begin{document}

\maketitle

\begin{abstract}
  Prior-Data Fitted networks (PFNs) have been very successful in tabular contexts, handling prediction tasks in context. However, they are designed for single-task inference, meaning that predicting several target values within a context requires repeated forward calls and precludes inter-task information sharing. We propose TabPFN-MT, which is trained on an expanded multi-target synthetic prior to capture inter-task dependencies in context. This model uses an expanded $y$-encoder and a shared decoder head to enable multitask in-context learning and simultaneous inference. The model is uniquely specialized for small-to-medium datasets by relying on in-context learning rather than traditional gradient-based training. Within this regime (averaging fewer than 1,000 samples), extensive evaluations across 344 datasets demonstrate that TabPFN-MT establishes a new state-of-the-art for deep tabular multitask learning. Furthermore, despite the inherent compute asymmetry of joint optimization, our model remains highly competitive with the latest state-of-the-art single-task ensembles. Notably, on multitask datasets it achieves an overall Accuracy rank of 4.89, the highest average rank among all models tested. Crucially, TabPFN-MT delivers this highly competitive performance while reducing the inference cost for $T$ tasks from $\mathcal{O}(T)$ to $\mathcal{O}(1)$ forward passes, offering a massive computational efficiency improvement for multi-target tabular applications.
\end{abstract}

\section{Introduction}
\label{sec:intro}

Tabular data is common across a wide range of domains, making it an important modality for the introduction of machine learning (ML) into real-world applications. In particular, there are many cases where datasets have several related outputs of interest. For instance, a medical trial may have multiple key readouts, or in finance, it may be valuable to jointly model an asset's return and volatility.

Despite the prevalence of multi-target problems, deep multitask learning (MTL) for tabular data has predominantly focused on massive datasets, such as e-commerce and recommender systems. In critical low-data regimes, however, single-task models like gradient-boosted decision trees (GBDTs) remain the standard. Applying these single-task architectures to multi-target problems requires fitting independent models for every target. This approach inherently prevents inter-task information sharing (precluding positive transfer) and multiplies the computational cost by the number of targets $T$.

Recently, TabPFN has emerged as a leading model in low-data tabular domains, leveraging in-context learning (ICL) to tackle small datasets with a higher-capacity model~\citep{hollmann_accurate_2025}. However, current tabular Prior-Data Fitted Networks (PFNs) are constrained to single-task inference. Thus, predicting  $T$ targets requires $T$ independent forward passes, leading to a massive waste of computation, re-calculating similar representations of the same tabular dataset. 

Simultaneously inferring multiple label distributions and modelling their interactions based on limited context is a substantially harder problem than standard scalar ICL. However, evidence from Natural Language Processing demonstrates that pretrained foundation models are capable of emergent multitask ICL~\citep{xiong_everything_2025}. We hypothesize that this capability can be translated to the tabular domain. Because a PFN can attend across the entire dataset context (both features and targets), it should be able to learn useful shared representations to support multiple simultaneous predictions. This approach requires only a single forward pass for multitask prediction, opening the possibility for positive transfer and improved predictions overall.

In this work, we introduce \textbf{TabPFN-MT}, the first tabular foundation model natively pretrained for multitask in-context learning. Our core contributions are:
\begin{enumerate}
    \item \textbf{Multitask Extension of PFN Paradigm:} This is the first MTL model to extend the PFN framework to tackle problems with a variable number of targets through joint inference. Crucially, this enables amortized inference: a single pretrained model can be applied to a wide range of datasets with varying feature and target dimensionality without dataset-specific training, instead leveraging ICL during inference. 
    \item \textbf{Multitask Architecture:} To support this flexibility, TabPFN-MT uses a dynamic $y$-encoder and a shared projection decoder. The encoder scales inputs depending on the number of tasks in a dataset and uses dynamic zero-padding to maintain consistent dimensions. The decoder generates outputs in a single inference pass, which are then sliced into per-task logit vectors.
    \item \textbf{Synthetic Prior Design} We demonstrate that pretraining this architecture on datasets generated by a single, complex Structural Causal Model (SCM) provides the necessary spectrum of task correlations to successfully teach a transformer to leverage shared representations in context. This prior is sampled symmetrically allowing for stable training without loss balancing techniques common in other multitask works.
    \item \textbf{Efficient State-of-the-Art Low-Data MTL:} Through extensive evaluation across 344 datasets in the small-to-medium data regime ($< 5,000$ samples), TabPFN-MT establishes a new state-of-the-art for tabular MTL. Furthermore, it achieves predictive performance competitive with compute-heavy single-task ensembles, while its architecture reduces computational overhead by reducing multi-target inference cost from $\mathcal{O}(T)$ to $\mathcal{O}(1)$ forward passes. 
\end{enumerate}

To support open science and future work building on these contributions, details about the future release of source code are included in \cref{app:reproducibility}.

The paper is organized as follows: \cref{sec:related-work} reviews prior work. \cref{sec:method} details the multitask prior and TabPFN-MT architecture. \cref{sec:experiment} presents our comprehensive evaluations and comparisons to baselines, and \cref{sec:discuss} concludes with limitations and future directions. 

\section{Related Work}
\label{sec:related-work}

\paragraph{Standard Tabular Models}
While deep learning has come to dominate ML applications across other modalities and domains, GBDTs remain the primary baseline for tabular data due to their strong handling of axis-aligned features~\citep{chen_xgboost_2016, ke_lightgbm_2017, prokhorenkova_catboost_2018}. However, their additive ensemble structure limits the applicability of multitask learning. GBDT methods tend to treat each task as a separate objective for optimization, using different trees or leaves for each target. This means that the models do not form shared representations across tasks and there is not a mechanism to share information across the targets. Additionally, while GBDT models tend to require fewer parameters and offer strong out-of-the-box baselines, achieving state-of-the-art performance still requires rigorous hyperparameter optimization (HPO)~\citep{mcelfresh_when_2023}.

\paragraph{Tabular Deep Learning}

While traditional ML approaches tend to operate on the raw data, models like SAINT~\citep{somepalli_saint_2021}, FT-Transformer~\citep{gorishniy_revisiting_2021}, and TabNet~\citep{arik_tabnet_2021} use attention mechanisms to create inter-feature representations. These models develop more rich internal representations but remain single task and single dataset; they must be trained from scratch for each new dataset and target.

There have been several works that apply MTL to large tabular datasets. Multi-Gate Mixture-of-Experts (MMoE)~\citep{ma_modeling_2018}, Progressive Layered Extraction (PLE)~\citep{tang_progressive_2020}, and Shared and Task-specific EMbeddings (STEM)~\citep{su_stem_2024} all use a mixture of experts (MoE) approach with variations to mitigate negative transfer between tasks. In contrast, MultiTab introduced a multitask masked attention mechanism to moderate the competition between tasks and improve multi-target performance~\citep{sinodinos_multitab_2026}. All of these MTL models are trained on the target dataset and the architectures must be fixed to the exact number of tasks seen during training. In contrast, this paper looks to develop a more flexible model that leverages in-context learning to infer task structure at inference time with less available data.

\paragraph{Tabular Prior-Data Fitted Networks}

In low-data regimes, PFNs have been very successful, pretraining on synthetic datasets and then adapting to target datasets through in-context learning~\citep{muller_transformers_2022}. TabPFN has achieved state-of-the-art performance across a range of tabular benchmarks following this paradigm~\citep{hollmann_tabpfn_2023, hollmann_accurate_2025, grinsztajn_tabpfn-25_2026}. Subsequent works have looked to improve PFN scalability via linear attention~\citep{zeng_tabflex_2025} and enhance pretraining with real-world data~\citep{ma_tabdpt_2025}.

All of these tabular PFNs have assumed single-task, scalar labels, an assumption that breaks down in multitask settings. Using a single-task model for multitask problems requires distinct inferences for each target, leading to redundant calculation and not allowing for shared information between the tasks. \citet{sinodinos_multitask-informed_2026} investigated adapting single-task TabPFN to a multitask setting via multitask domain-specific fine-tuning. In this work, we take a complementary approach, investigating whether a PFN can learn multitask in-context inference through natively multitask pretraining with a modified prior.

\section{Methodology}
\label{sec:method}

\begin{figure}
    \centering
    \begin{subfigure}[b]{0.95\textwidth}
        \centering
        \includegraphics[width=\textwidth]{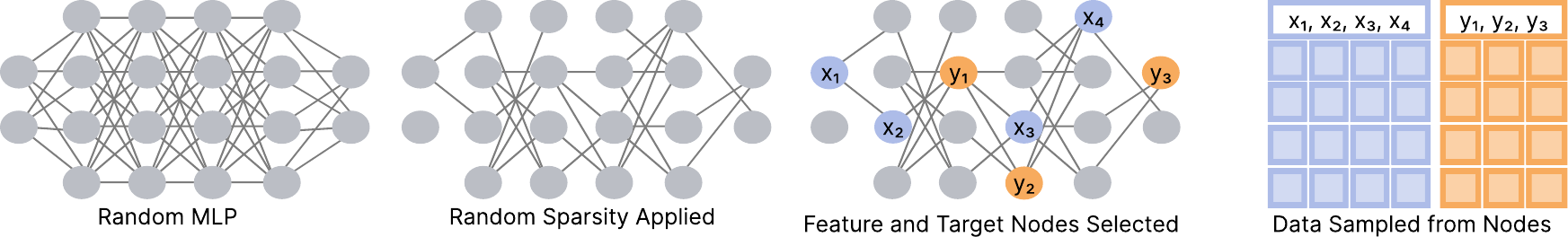}
        \caption{Synthetic multi-target data generation from a single underlying Structural Causal Model (SCM).}
        \label{fig:data_gen}
    \end{subfigure}
    \\[0.5em]
    \begin{subfigure}[b]{0.95\textwidth}
        \centering
        \includegraphics[width=\textwidth]{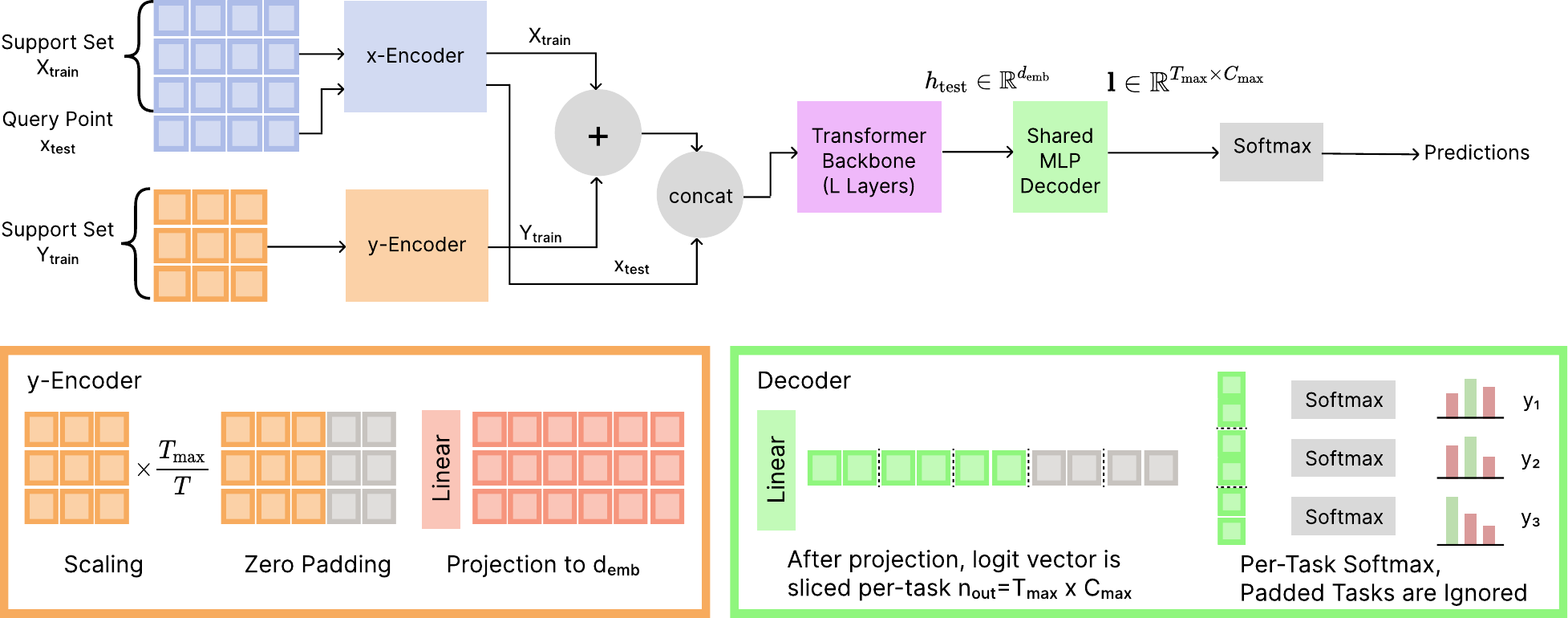}
        \caption{The TabPFN-MT architecture, with an expanded y-encoder and decoder to handle multiple targets.}
        \label{fig:mtl_model}
    \end{subfigure}
    \caption{\textbf{The TabPFN-MT Framework.} The model relies on a synthetic multitask prior generated via highly sparse Multi-Layer Perceptrons (MLPs) simulating causal directed acyclic graphs (DAGs). During inference, the architecture leverages an expanded $y$-encoder and a dynamically sliced decoder head to perform simultaneous multitask in-context learning over a shared Transformer backbone.}
    \label{fig:block-diagram}
\end{figure}

An overview of the complete TabPFN-MT framework, detailing both the synthetic data generation process and the model architecture, is illustrated in \cref{fig:block-diagram}.

\subsection{Multitask In-Context Learning}
\label{sec:mtl-icl}

\citet{muller_transformers_2022} established that PFNs make predictions in context via approximate Bayesian inference. That is, given a labelled training dataset $D_{\text{train}}$ and a test point $x_{\text{test}}$, the pretrained PFN $q_\theta$ approximates the Posterior Predictive Distribution (PPD) for the label, $p(y_{\text{test}}\mid x_{\text{test}},D_{\text{train}})$. In prior PFN models, the model predicts a single scalar target per query, $q_\theta(y_{\text{test}} \mid x_{\text{test}},D_{\text{train}}) \approx p(y_{\text{test}}\mid x_{\text{test}},D_{\text{train}})$
~\citep{hollmann_tabpfn_2023, hollmann_accurate_2025}.  In our multitask extension, the same PPD approximation underlies the model, but the context is expanded such that $D_{\text{train}} = \{(x_i, \mathbf{y}_i)\}_{i=1}^N$. The target becomes a vector $\mathbf{y}_i=(y^{(1)}, y^{(2)}, \dots, y^{(T)})$ corresponding to the $T$ distinct tasks present in a given dataset $D_{\text{train}}$. Our model prioritizes simultaneous inference for all targets, thus the model approximates the joint distribution of these targets as conditionally independent given the shared network representation. %\r{$D$}

Let $\mathbf{h}_{\text{test}}$ denote the final hidden state of the query point output by the transformer backbone, summarizing the labelled context $D_{\text{train}}$ and the query features $x_{\text{test}}$. The shared Multi-Layer Perceptron (MLP) decoder maps this representation to a joint logit vector $\mathbf{l} = \text{MLP}_{\theta_{\text{head}}}(\mathbf{h}_{\text{test}})$.

The information sharing between tasks occurs natively within the hidden layers of this shared MLP, allowing the network to leverage learned correlations between tasks to shape the joint logit vector. The static output $\mathbf{l}$ is then dynamically sliced into task-specific logit vectors $\mathbf{l}^{(t)} \in \mathbb{R}^{C_{\text{max}}}$, where $C_{\text{max}}$ is the maximum number of classes per task. The joint PPD is thus factorized at the output layer, where each marginal probability is parameterized by its respective logit slice:
\begin{equation}
    q_\theta(\mathbf{y}_{\text{test}} \mid x_{\text{test}}, D_{\text{train}}) = \prod_{t=1}^{T} q_\theta(y_{\text{test}}^{(t)} \mid \mathbf{l}^{(t)}) = \prod_{t=1}^{T} \text{Softmax}(\mathbf{l}^{(t)})_{y_{\text{test}}^{(t)}}.
\end{equation}

\subsection{Multitask Structural Causal Model Prior}

To generate the synthetic datasets used during the prior-fitting phase, we build upon the SCM prior introduced in the original TabPFN~\citep{hollmann_tabpfn_2023, muller_transformers_2022}. 
Datasets are generated by sampling a random directed acyclic graph (DAG) representing an SCM, propagating noise variables through non-linear deterministic functions, and selecting a subset of nodes to act as observed features $X$ and a single target $Y$. The dataset is then partitioned into training $D_{\text{train}}=(X_{\text{train}},Y_{\text{train}})$ and test sets $D_{\text{test}}=(X_{\text{test}},Y_{\text{test}})$. This approach yields diverse datasets with complex, conditionally dependent features driven by forward and backward causation. We adopt this foundation but extend the data-generation process to simulate multi-target tabular environments. Instead of a scalar target, the multitask SCM must sample a multi-target matrix $\mathbf{Y}$.

A critical requirement for a robust multitask prior is the ability to generate datasets with varying degrees of inter-task correlation. If tasks are perfectly correlated, multitask learning is trivial; if they are entirely independent, models are susceptible to negative transfer. In practice, these SCMs are parameterized as MLPs, where causal DAG structures are simulated by applying high sparsity masks to the network weights (\cref{fig:data_gen}). The complexity of the task relationships is modulated by varying the depth, width, sparsity bounds, and the number of underlying MLPs sampled per dataset. After testing multiple configurations for sampling multiple targets (detailed in \cref{app:prior-details}), we found that using a single, shared, high-complexity SCM for the prior yielded the best downstream model.

Using a large underlying SCM means that in some cases features and targets will be sampled close together (yielding high relatedness) whereas in other cases they will be very distinct (yielding low relatedness). Intuitively, this variation should lead to a model which is robust to different strengths of task relationships. In all cases, there does exist a shared underlying causal structure between all features and targets which should incentivize the model to form shared representations to capture positive transfer. In this work, we restrict our focus to multi-target classification (binary and multiclass) supporting classification up to 10 classes per target. We leave heterogeneous combinations of continuous and categorical targets as an interesting direction for future work. 

\subsection{Multitask Transformer}

Following the original TabPFN~\citep{hollmann_tabpfn_2023}, our model processes tabular data as a set of rows, utilizing self-attention to route information between the labelled context $D_{\text{train}}$ and unlabelled query $x_{\text{test}}$. To support multitask in-context learning, the transformer backbone remains unmodified; all architectural changes are isolated to the boundaries of the network: the input $y$-encoder and the output decoder head. During inference, we similarly adopt their 32-permutation ensembling strategy to reduce positional bias (see \cref{app:ensemble} for details).

\subsubsection{Expanded \texorpdfstring{$y$}{y}-Encoder}

In a standard TabPFN, the target encoder projects a scalar label into the shared embedding space. To accommodate our multitask formulation, we expand this to ingest the full target vector $\mathbf{y}_i$ (\cref{fig:mtl_model}). A primary challenge of in-context multitask learning is handling datasets with a variable number of tasks $T$ up to an architectural maximum $T_{\text{max}}$.

To resolve this without altering the underlying network dimensions, our $y$-encoder employs a fixed-width base projection preceded by a dynamic scaling and padding module. Given an input target vector of length $T \leq T_{\text{max}}$, the vector is first multiplied by a magnitude-preserving scaling factor ($T_{\text{max}}/T$). This critical scaling step ensures that the overall variance and magnitude of the network activations remain stable, regardless of how many tasks are present in the context. The scaled vector is subsequently zero-padded to the fixed maximum dimension $T_{\text{max}}$ before being projected into the model's $d_{\text{emb}}$-dimensional working space.

\subsubsection{Expanded Decoder Head}

Similarly, the decoder head must be expanded to output predictions for all potential tasks simultaneously. We utilize a single, shared Multi-Layer Perceptron (MLP). The decoder consists of two linear layers separated by a GELU activation, projecting the final hidden representation of the query point $\mathbf{h}_{\text{test}}$ through a hidden dimension $d_{\text{hid}}$ to a static output dimension $n_{\text{out}}$. 

Crucially, this decoder design is chosen instead of other paradigms common in MTL models such as task-specific routing or autoregressive decoding. The introduction of task-specific heads would impose an assumption of consistent dataset structure, with a constant number of targets and systematic relationships between them. In our ICL paradigm, the architecture must accommodate arbitrary task dimensionality and different inter-task relationships, thus there is no basis to train task-specific heads. Autoregressive decoding would allow for a variable numbers of targets but would introduce an artificial task ordering and degrade inference to $\mathcal{O}(T)$.

In our configuration, $n_{\text{out}}$ is defined by the product of the architectural maximums: $T_{\text{max}} \times C_{\text{max}}$, where $C_{\text{max}}$ is the maximum number of classes per task. For example, with $T_{\text{max}}=5$ and $C_{\text{max}}=10$, the decoder maps to a 50-dimensional logit vector. As formalized in \cref{sec:mtl-icl}, this static output vector is then dynamically sliced during the loss calculation to isolate the specific logits for the $T$ valid tasks present in that dataset, while ignoring the outputs corresponding to the padded dimensions. These maximums are critical in maintaining strict dimensional consistency that enables efficient, parallelized, batched pretraining across synthetic datasets with different structures.

\subsection{Training Objective and Configuration}
\label{sec:model-and-training}

The training objective remains aligned with prior PFNs: minimizing the cross-entropy on held-out test sets $D_{\text{test}} \subset D$.
% $(X_{\text{test}}, Y_{\text{test}}) \in D$. 
To accommodate varying numbers of tasks per dataset, the model dynamically slices the output dimensions to match the target. The loss is calculated independently only for the $T$ valid tasks present in the current dataset context. To ensure training stability across different datasets, the total loss is the average of the negative log-likelihoods across these valid tasks for all test points:
\begin{equation}
    \mathcal{L}_{\text{PFN}}(\theta) = \mathbb{E}_{D \sim p(D)} \mathbb{E}_{(x_\text{test},\mathbf{y}_{\text{test}})\sim D_\text{test}} \left[ \frac{1}{T} \sum_{t=1}^{T} - \log q_\theta(y_{\text{test}}^{(t)} \mid \mathbf{l}^{(t)}) \right].
\end{equation}

In our synthetic multi-target training setup, there is no need for gradient interventions which are common in MTL deep learning~\citep{cipolla_multi-task_2018, chen_gradnorm_2018, yu_gradient_2020}. All targets are sampled symmetrically thus there are no systemic differences in task scale, noise, or difficulty across the synthetic datasets. Thus, simple uniform averaging of cross-entropies is sufficient for stable training. 

Among nine tested configurations varying embedding dimension and number of layers, the best-performing backbone had just 8.1M parameters; notably fewer than TabPFN v1's 25.8M~\citep{hollmann_tabpfn_2023}. This compression suggests training may be approaching a \emph{complexity ceiling} of the synthetic prior, and the reduced parameter count directly benefits inference speed and memory footprint, furthering the efficiency gains discussed in \cref{sec:comp-scaling}. Full configuration details are in \cref{app:model-scaling}.

TabPFN-MT is trained for 200 epochs on 3.28M synthetic datasets using Adam~\citep{Kingma2014AdamAM}; complete training details are provided in \cref{app:training-details}.

\section{Experiments}
\label{sec:experiment}

In this section, we evaluate the predictive performance and computational efficiency of TabPFN-MT. We first detail the datasets, baselines, and metrics used in our evaluation, followed by a comprehensive analysis of the results.

\subsection{Experimental Setup}

To ensure a rigorous and fair comparison, our evaluation framework encompasses a wide variety of tabular environments and state-of-the-art models. The specific details of our setup are outlined below.

\subsubsection{Real-world Datasets}

To comprehensively evaluate TabPFN-MT across diverse data regimes, we curated a benchmark of 344 real-world datasets across four distinct categories. First, to establish a Native Multitask baseline, we curated multi-target classification datasets from OpenML, filtering out instances of extreme class imbalance (Imbalance Ratio $> 16$) and restricting bounds to a maximum of 5,000 samples, 100 features, and 5 targets. Second, we utilized a subset of Standard Single-Task ($T=1$) OpenML datasets drawn from the validation suite established by \citet{hollmann_tabpfn_2023}. Third, to robustly evaluate performance across varying target correlations, we constructed Feature-Derived Multitask datasets by programmatically repurposing input features as new joint target vectors from established single-target benchmarks. Finally, we evaluated two Subsampled Large-Scale datasets, AliExpress~\citep{li_improving_2020, xi_modeling_2021} and ACS Income~\citep{ding_retiring_2021, ma_modeling_2018}, randomly subsampled down to $\{500, 1000, 2500, 5000\}$ instances across five seeds to adapt these massive benchmarks to the low-data regime targeted by our model. Summary statistics and implementation details for all categories of datasets are provided in \cref{app:dataset-details}.

\subsubsection{Baselines}

\paragraph{Single-Task Models}

To fully contextualize the performance of TabPFN-MT, we compare it against state-of-the-art multitask models as well as independent ensembles of single-task models. For the single-task baselines, we train a separate model for each target, a standard practice in MTL~\citep{goodfellow_deep_2016}. For our single-task baselines, we evaluate a standard MLP, alongside Random Forest (RF)~\citep{breiman_random_2001}, XGBoost~\citep{chen_xgboost_2016}, and CatBoost~\citep{prokhorenkova_catboost_2018}, as gradient-boosted decision trees have long been held as the strongest defaults for tabular data. For deep single-task baselines, we include TabTransformer (Tab-T)~\citep{huang_tabtransformer_2020}, FT-Transformer (FT-T)~\citep{gorishniy_revisiting_2021}, and SAINT~\citep{somepalli_saint_2021}. We also evaluate against TabPFN v1~\citep{hollmann_tabpfn_2023}, whose architecture TabPFN-MT is based upon, and the most recent TabPFN release, v2.6, which builds upon the changes described in the TabPFN v2.5 report~\citep{grinsztajn_tabpfn-25_2026}.

% Note that single-task baselines inherently incur a computational penalty that scales linearly with the number of tasks, as they require independent training and inference passes per target, unlike the joint optimization of multitask architectures, this is explored in \cref{sec:comp-scaling}.

\paragraph{Multitask Models}

For multitask baselines, we compare against large-scale deep models, including MoE MLP architectures—MMOE~\citep{ma_modeling_2018}, PLE~\citep{tang_progressive_2020}, and STEM~\citep{su_stem_2024}—as well as the transformer-based MultiTab (MTT)~\citep{sinodinos_multitab_2026}. It should be noted that the small dataset sizes that this paper examines are outside of the regime that these models were developed for.

To ensure a rigorous comparison, all baselines undergo hyperparameter optimization (HPO). Details on baseline implementations and the search spaces and HPO processes are detailed in \cref{app:baseline-details}. 

\subsubsection{Metrics}

To evaluate classification performance, we report Accuracy, F1 Score, and Area Under the Receiver Operating Characteristic Curve (ROC AUC) aggregated across all $T$ targets. Since our model encounters both binary and multiclass classification problems, we use both binary and multiclass formulations of those metrics. 

To assess benefits of multitask performance for a model $m$, we use multitask gain ($\Delta_m$) from \citet{maninis_attentive_2019}. $\Delta_m$ is calculated as the average change in performance for a measure $M$ compared to a single-task baseline $b$ across all tasks $i=1,\dots,T$. For the baseline we use a single-task MLP, 

\begin{equation}
    \Delta_m=\frac{1}{T}\sum_{i=1}^T\frac{M_{m,i}-M_{b,i}}{M_{b,i}} .
\end{equation}

Because our metrics are strictly ``higher is better,'' a positive $\Delta_m$ indicates that the multitask model outperforms the single-task baseline.

Detailed formulations on multiclass metric handling, our cross-validation setup, and our statistical significance testing pipeline (Friedman and Nemenyi tests) are provided in \cref{app:extended-experiments}.

\subsection{Predictive Performance}

\begin{figure}[t]
    \centering
    \includegraphics[width=\linewidth]{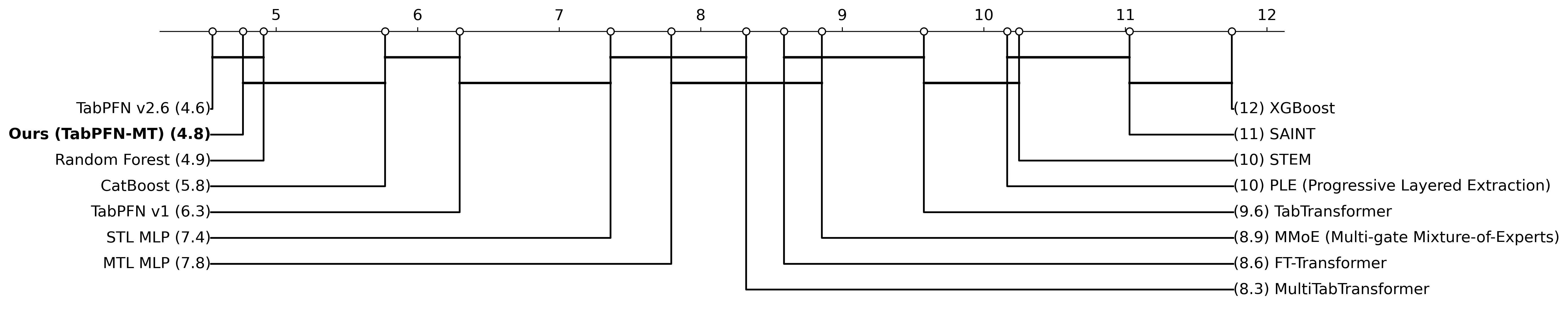}
    \caption{TabPFN-MT is in the top clique for Accuracy and is the highest-ranked multitask model. The diagram displays the average rank of each model across all 344 datasets evaluated, with lower ranks indicating better performance (further left). Thick horizontal lines connect groups of models (cliques) that do not exhibit statistically significant performance differences from one another based on a Nemenyi post-hoc test ($p < 0.05$). While TabPFN v2.6 achieves the best overall average rank, our proposed TabPFN-MT performs competitively within the same top-tier statistical clique and significantly outperforms all other multitask baselines.}
    \label{fig:cd-accuracy}
\end{figure}

TabPFN-MT ranks first among all evaluated multitask models across every metric and closely approaches the performance of state-of-the-art single-task ensembles (\cref{tab:summary_adjusted_all_models}). The Critical Difference (CD) diagram for Accuracy (\cref{fig:cd-accuracy}) places our model in the top-performing clique, indicating statistical equivalence with the strongest single-task baselines. Crucially, TabPFN-MT is the \textit{only} multitask model in this top clique, achieving this performance with $\mathcal{O}(1)$ computational complexity regarding the number of tasks, compared to the $\mathcal{O}(T)$ scaling required by the single-task baselines (\cref{tab:timing_t1_t5_scaling}). Additional CD diagrams in \cref{app:extended-results} confirm this trend: TabPFN-MT remains the highest-ranked MTL model overall, belonging to the top clique for both F1 and ROC AUC.

To better characterize where TabPFN-MT excels, we stratify predictive performance by dataset origin (detailed in \cref{app:extended-results}). We observe that while deep MTL baselines (e.g., Multitask MLP and MultiTab) show relative improvement on the subsampled Large MTL datasets due to the higher sample counts, they still fail to outperform strong single-task models even in this regime. In contrast, TabPFN-MT demonstrates exceptional performance on the Derived MTL datasets, achieving the highest overall Accuracy (0.526) and outperforming all other multitask baselines by a wide margin across all metrics. Because these datasets are generated by repurposing existing input features as joint targets, they inherently possess strong, structured causal relationships. This environment closely aligns with the inductive bias of our SCM synthetic prior, where multiple targets are sampled from the same dense causal graph. Consequently, TabPFN-MT is highly effective at identifying these inter-task dependencies in context and leveraging them for positive cross-task transfer.

Conversely, joint inference introduces predictive trade-offs, most notably reflected in the macro-averaged F1 scores. While TabPFN-MT consistently leads the multitask baselines, it occasionally trails the absolute best single-task models (specifically TabPFN v2.6 and Random Forest) in F1 performance, particularly on the Native MTL and Large MTL splits. Single-task architectures dedicate their entire functional capacity to optimizing a single, isolated decision boundary, allowing them to more precisely capture minority classes in imbalanced scenarios. In contrast, TabPFN-MT must distribute its shared representation and decoder capacity across multiple targets simultaneously (explored in \cref{app:pred-vs-capacity}). As observed in our target-scaling ablation, this shared capacity creates an information bottleneck that can slightly degrade performance on complex, heavily imbalanced, or weakly correlated tasks.

\begin{table}[ht]
\small
\centering
\caption{Overall comparison across key metrics with all models included. Top-level columns separate Single-Target and Multi-Target dataset evaluation views. Within each column, \textbf{bold} marks the best MTL model value and \underline{underlined} text marks the best STL model value (ties share the same formatting).}
\label{tab:summary_adjusted_all_models}
\setlength{\tabcolsep}{5pt}
\begin{tabular}{lrrrrrr}
\toprule
 & \multicolumn{3}{c}{Single-Target Datasets 
% (N=61)
} & \multicolumn{3}{c}{Multi-Target Datasets 
% (N=283)
} \\
\cmidrule(lr){2-4} \cmidrule(lr){5-7}
Model & AUC & Acc. & F1 & AUC & Acc. & F1 \\
\midrule
\multicolumn{7}{l}{\fbox{Multitask Models:}} \\
TabPFN-MT (Ours) & \textbf{0.826} & \textbf{0.805} & \textbf{0.674} & \textbf{0.727} & \textbf{0.578} & \textbf{0.421} \\
MT-MLP & 0.790 & 0.741 & 0.545 & 0.683 & 0.518 & 0.344 \\
MMoE & 0.769 & 0.727 & 0.519 & 0.675 & 0.505 & 0.324 \\
PLE & 0.751 & 0.720 & 0.481 & 0.652 & 0.486 & 0.293 \\
STEM & 0.737 & 0.708 & 0.478 & 0.652 & 0.486 & 0.292 \\
MTab & 0.806 & 0.774 & 0.630 & 0.665 & 0.515 & 0.331 \\
\multicolumn{7}{l}{\fbox{Single-task Models:}} \\
ST-MLP & 0.786 & 0.765 & 0.658 & 0.677 & 0.536 & 0.398 \\
RF & 0.852 & 0.816 & 0.691 & 0.708 & \underline{0.565} & \underline{0.432} \\
XGB & 0.768 & 0.694 & 0.396 & 0.662 & 0.459 & 0.265 \\
CatB & 0.848 & 0.810 & 0.679 & 0.709 & 0.544 & 0.418 \\
Tab-T & 0.761 & 0.738 & 0.564 & 0.649 & 0.502 & 0.335 \\
FT-T & 0.806 & 0.774 & 0.630 & 0.676 & 0.516 & 0.335 \\
SAINT & 0.732 & 0.727 & 0.518 & 0.632 & 0.483 & 0.278 \\
TabPFN-v1 & 0.834 & 0.805 & 0.665 & 0.720 & 0.527 & 0.384 \\
TabPFN-v2.6 & \underline{0.877} & \underline{0.849} & \underline{0.730} & \underline{0.734} & 0.562 & 0.417 \\
\bottomrule
\end{tabular}
\end{table}

\subsection{Computational Efficiency}
\label{sec:comp-scaling}

To systematically evaluate the behaviour of TabPFN-MT as the target space grows, we benchmark the total floating-point operations (FLOPs) required across increasing numbers of tasks. While comparing FLOPs provides a rigorous, hardware-agnostic measure of computational cost, tree-based models like GBDT baselines cannot be captured in this analysis. Detailed information regarding the FLOP counting methodology and wall-clock comparisons that include GBDT baselines are provided in \cref{app:extended-efficiency}.

We use the MultiTab framework~\citep{sinodinos_multitab_2026} to generate synthetic tabular datasets. To benchmark computational cost, datasets are standardized to 1,000 samples, 50 features, a noise level of 0.01, and polynomial degrees ranging from 1 to 3. We vary the number of targets from 1 to 5 and fix task correlation at 0.4.

\begin{figure}
    \centering
    \includegraphics[width=0.95\linewidth]{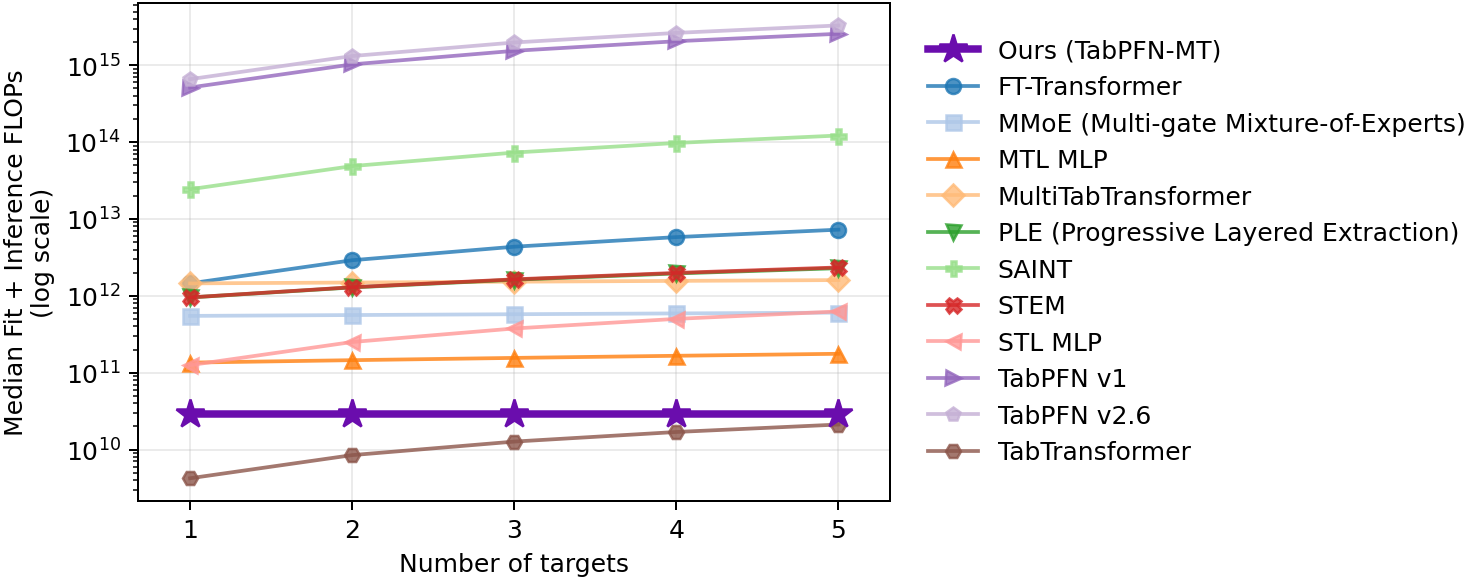}
    \caption{Computational cost scaling across architectures. Total FLOPs (log scale) from training and inference are evaluated on synthetic datasets containing 1 to 5 targets. Single-target architectures exhibit $\mathcal{O}(T)$ scaling, requiring linearly more compute for each additional target. In contrast, our proposed model, TabPFN-MT (solid purple line), natively infers all targets in a single forward pass, achieving $\mathcal{O}(1)$ scaling and maintaining constant computational efficiency. Note that this measurement excludes the cost of HPO which is relevant for all models except for the PFNs.} 
    \label{fig:flops-scaling} 
\end{figure}

As illustrated in \cref{fig:flops-scaling}, single-target baselines (e.g., standard TabPFN variants, SAINT, and FT-Transformer) incur a linear computational cost ($\mathcal{O}(T)$) as the number of targets increases. These models require independent training and inference processes for each target. In contrast, multitask models like TabPFN-MT handle targets simultaneously allowing for constant computational cost ($\mathcal{O}(1)$)  even as the number of targets increases.

Furthermore, because TabPFN-MT extends the PFN paradigm, it does not require dataset-specific training, achieving highly competitive computational costs that fall below other multitask baselines. The lightweight model backbone described above in \cref{sec:model-and-training} also means that TabPFN-MT's cost is orders of magnitude below single-task TabPFNs (even in the single-target domain).

It is important to note that this dataset-level analysis does not capture the initial pretraining phase of TabPFN-MT, which required approximately 4 hours across eight RTX 5000 GPUs. However, unlike traditional baselines that require costly hyperparameter optimization and iterative training for every newly encountered dataset, TabPFN-MT adapts entirely in context. Consequently, this one-time pretraining investment is quickly amortized across downstream applications, ultimately providing a massive reduction in cumulative compute for multi-target tabular environments.

\section{Discussion and Conclusion}
\label{sec:discuss}

In this work we introduce TabPFN-MT, a novel tabular foundation model that establishes a new state-of-the-art for deep tabular models in the small-to-medium sized data regime. On the datasets tested, our work consistently outperforms existing MTL models and achieves predictive parity (Accuracy, F1, ROC AUC) with top-performing ensembles of single-task models, including TabPFN v2.6 and GBDTs. Crucially, it achieves this level of performance while reducing the multi-target inference cost from $\mathcal{O}(T)$ to $\mathcal{O}(1)$ compared to the application of target-specific single-task models. Furthermore, TabPFN-MT uses in-context adaptation rather than gradient-based for dataset-specific adaptation, thus avoiding the negative transfer risk which is a problem in traditional MTL approaches. The pretraining phase uses synthetic data which does not have consistent gradient conflicts. During the forward pass, the transformer architecture has sufficient capacity in the multi-head attention mechanism to dynamically attend to features and labels in the support set, avoiding destructive interference but still offering the potential for positive cross-task transfer.

\paragraph{Limitations} While foundation models inherently require an initial pretraining investment, TabPFN-MT trains in just four hours across eight GPUs. This represents a fraction of the compute typically required for foundation models. This one-time cost is amortized across a wide range of downstream applications which offer high performance without dataset-specific finetuning or hyperparameter optimization. 

Currently the model is scoped for classification problems with a maximum of 5 tasks and 10 classes but the same approach could be extended to regression following the approach with a piece-wise distribution taken by \citet{hollmann_accurate_2025}. The task and class count limits could be scaled in future if motivated by downstream applications. 

Furthermore, our work finds that a low-parameter (8.1M) model variant can outperform larger models. This suggests that this model had sufficient capacity to map the prior's complexity. This suggests that further development of this class of models will not come from simply adding layers, but instead from engineering more complex proxy distributions which still transfer to real-world data. This finding is aligned with recent work that has begun to introduce real-world data into the pretraining of tabular foundation models like \citet{ma_tabdpt_2025} and \citet{grinsztajn_tabpfn-25_2026}.

\paragraph{Societal Impact} TabPFN-MT does not require the same compute-expensive hyperparameter optimization phase as other tabular methods; thus its adoption could help reduce environmental impact of tabular deep learning. As tabular data is frequently found in high-sensitivity domains (e.g., healthcare, finance), practitioners should be mindful of historical biases in the data used for predictions. Because the model uses in-context adaptation to make predictions based on limited data, it will propagate biases captured in a target dataset into predictions. The rigorous curation of unbiased support sets is crucial to ensure responsible and equitable outcomes when applying this work.

\s{\begin{ack}
The authors gratefully acknowledge the financial and computational support provided by the iSMART lab at McGill University.
\end{ack}}

\bibliographystyle{plainnat}
\bibliography{MTL_TabPFN}

%%%%%%%%%%%%%%%%%%%%%%%%%%%%%%%%%%%%%%%%%%%%%%%%%%%%%%%%%%%%

\appendix

\section{Reproducibility Statement}
\label{app:reproducibility}

To ensure the reproducibility of our results, we will provide source code, hyperparameter configurations, and instructions for generating the synthetic data in the future upon the formal publication of this work. We emphasize that this approach is accessible to the broader academic community. A complete pretraining run of the multitask model takes just over four hours on a single node equipped with eight GPUs (NVIDIA A5000). 

\section{Multitask Prior Details}
\label{app:prior-details}

To systematically determine the optimal synthetic data generation strategy, we employed a multi-stage Bayesian hyperparameter optimization pipeline using BOHB (Bayesian Optimization and Hyperband) through Weights and Biases. This search was conducted in three successive phases to isolate the effects of prior structure and training dynamics. 

First, we conducted a broad exploratory sweep over the structural paradigms, allowing the prior to sample varying numbers of underlying SCMs to map the relationship between task correlation and model capacity (\cref{tab:sweep_bounds:mtl_prior_design_bayes_hyperband}). Second, having identified promising prior configurations, we performed a dedicated search over the transformer's training hyperparameters to ensure stable optimization during joint inference (\cref{tab:sweep_bounds:mtl_train_config_bayes_hyperband}). Finally, guided by our empirical findings that fewer SCMs yield superior downstream performance, we executed a highly focused search constraining the prior strictly to a single, high-complexity SCM to fine-tune the causal graph parameters (\cref{tab:sweep_bounds:mtl_prior_design_num_mlps_1_bayes_hyperband}).

\begin{table}[t]
\centering
\small
\setlength{\tabcolsep}{12pt}
\caption{Search bounds for initial sweep over prior meta-hyperparamters with multiple underlying SCMs.}
\label{tab:sweep_bounds:mtl_prior_design_bayes_hyperband}
\begin{tabular}{llcc}
\toprule
 & & \multicolumn{2}{c}{Search Space} \\
\cmidrule(lr){3-4}
Hyperparameter & Distribution & Lower Bound & Upper Bound \\
\midrule
Number of MLPs & Integer Uniform & [1, 5] & [2, 5] \\
Number of layers & Integer Uniform & [2, 6] & [8, 20] \\
MLP hidden dimension & Integer Uniform & [4, 64] & [96, 320] \\
Noise standard deviation & Log Uniform & [1.00e-04, 0.03] & [0.08, 0.8] \\
Categorical feature prob. & Uniform & \multicolumn{2}{c}{[0, 1]} \\
\bottomrule
\end{tabular}
\end{table}

\begin{table}[t]
\centering
\small
\setlength{\tabcolsep}{12pt}
\caption{Search bounds for secondary sweep over prior meta-hyperparamters with single SCMs.}
\label{tab:sweep_bounds:mtl_prior_design_num_mlps_1_bayes_hyperband}
\begin{tabular}{llcc}
\toprule
 & & \multicolumn{2}{c}{Search Space} \\
\cmidrule(lr){3-4}
Hyperparameter & Distribution & Lower Bound & Upper Bound \\
\midrule
Number of layers & Integer Uniform & [2, 6] & [8, 20] \\
MLP hidden dimension & Integer Uniform & [4, 64] & [96, 320] \\
Noise standard deviation & Log Uniform & [1.00e-04, 0.03] & [0.08, 0.8] \\
Categorical feature prob. & Uniform & \multicolumn{2}{c}{[0, 1]} \\
\bottomrule
\end{tabular}
\end{table}

\begin{table}[t]
\centering
\small
\setlength{\tabcolsep}{12pt}
\caption{Search bounds for training hyperparameter sweep.}
\label{tab:sweep_bounds:mtl_train_config_bayes_hyperband}
\begin{tabular}{lll}
\toprule
Hyperparameter & Distribution & Search Space \\
\midrule
Learning rate & Categorical & [null, 0.003, 0.001, 3.00e-04, 1.00e-04] \\
Warmup epochs & Categorical & [0, 5, 10, 20] \\
Gradient clip norm & Categorical & [0.5, 1, 2] \\
\bottomrule
\end{tabular}
\end{table}

To model this spectrum of task relatedness, we initially formalized the data-generating process under three distinct structural paradigms:
\begin{itemize}
    \item \textbf{Single Shared SCM:} A single causal graph is sampled. All $T$ targets and all features are drawn from the observed nodes of this unified graph, simulating domains with high latent feature sharing and highly correlated targets.
    \item \textbf{$T$ Distinct SCMs:} $T$ independent SCMs are sampled. Each target $y^{(t)}$ is drawn from its own distinct graph, while the feature set $X$ is constructed by pooling observed nodes across all $T$ graphs. This simulates distinct tasks that coincidentally share a feature space, yielding low inter-task correlation.
    \item \textbf{Distributed $n$-SCMs:} A hybrid approach where $n \sim U(1, T)$ SCMs are sampled. The features and targets are randomly distributed across these $n$ underlying graphs, creating clusters of related tasks within the same dataset.
\end{itemize}

To evaluate the proposed structural paradigms, we conducted an extensive hyperparameter search over the prior configurations. We initially hypothesized that sampling from a higher number of distinct SCMs ($n \gg 1$) would be necessary to generate sufficiently diverse, low-correlation tasks. However, our empirical evaluations demonstrated a strong negative correlation between the number of underlying MLPs and overall multitask performance.

The experiments revealed two primary failure modes when scaling the number of distinct SCMs. First, forcing the prior to utilize multiple distinct MLPs led to severe computational bottlenecks during synthetic data generation; the complex graph construction and sampling processes starved the GPU, significantly reducing training efficiency. Second, and more critically, datasets generated from many independent SCMs induced negative transfer during the transformer's in-context learning phase, degrading the model's ability to form unified latent representations.

Analysis of the top-performing configurations showed that the model heavily favours a minimal number of shared SCMs with higher individual complexity. Specifically, the optimal bounds restricted the prior to sampling between one and two underlying MLPs ($n \in [1, 2]$). 

Subsequent evaluation of models trained with a prior with a single shared SCM revealed that it further improved model performance. This is in line with the trend that priors with fewer underlying MLPs lead to better performance. Interestingly, our HP sweep revealed that priors that favour with more complex sparse SCMs (ie., more layers and higher hidden dimensions) than the parameters used in the original TabPFN work~\citep{hollmann_tabpfn_2023}. However, especially in the case of the number of layers, there are diminishing returns: increasing the number of layers does not monotonically increase resulting performance and increases the computational resources needed to generate synthetic data. Intuitively, this prior configuration incentivizes modelling of a single underlying causal structure, although different features and targets may be sampled from very different parts of the structure. The complete configuration of the top-performing prior is included in \cref{tab:prior_config_from_base_swift_sweep_7}.

The final resolved configuration for our top-performing prior (detailed in \cref{tab:prior_config_from_base_swift_sweep_7}) reflects these structural insights. To generate a sufficiently rich and varied multitask environment from a single causal graph, the prior relies on relatively deep (five to eigth layers) and wide (up to 278 hidden dimensions) MLPs. Crucially, to ensure these networks simulate sparse causal DAGs rather than dense neural representations, the configuration employs heavy regularization, including high MLP dropout probabilities and block-wise dropout. The injection of diverse activation functions (e.g., ELU, Tanh, Leaky ReLU) further increases the non-linear complexity of the generated functions. By sampling datasets with a dynamically varying number of targets ($T \in [1, 5]$) from this single robust structure, the resulting prior forces the transformer to learn flexible, shared representations that generalize effectively to real-world multitask tabular data.

\begin{table}[t]
\centering
\small
\setlength{\tabcolsep}{4pt}
\caption{Resolved prior configuration used for top-performing training run.}
\label{tab:prior_config_from_base_swift_sweep_7}
\begin{tabular}{llp{6.5cm}}
\toprule
Hyperparameter & Distribution & Definition \\
\midrule
Number of layers & TNLU & low=5, high=8, min=2, int \\
MLP hidden dimension & TNLU & low=57, high=278, min=4, int \\
Noise standard deviation & TNLU & low=6.94e-04, high=0.262207, min=0 \\
Categorical feature probability & Fixed & 0.287047 \\
Number of features & TNLU & low=1, high=100, min=10, max=100 \\
Number of outputs & Uniform & low=1, high=5, int \\
MLP dropout probability & Beta Meta Uniform & alpha\_low=0.1, alpha\_high=5, beta\_low=0.1, beta\_high=5, scale=0.9 \\
Init std & TNLU & low=0.1, high=5, min=0.01 \\
Number of causes & TNLU & low=1, high=12, min=1, int \\
MLP activations & Choice & [elu, tanh, identity, leaky\_relu] \\
Block-wise dropout & Choice & [true, false] \\
Random feature rotation & Choice & [true, false] \\
Is causal & Choice & [true, false] \\
Sampling & Choice & [normal, mixed] \\
\bottomrule
\end{tabular}
\end{table}

\section{Ensembling}
\label{app:ensemble}

Following the methodology of \citet{hollmann_tabpfn_2023}, we employ an ensembling technique at inference time to improve predictive robustness and mitigate positional bias. Rather than relying on a single forward pass, we average the model's predictions over an ensemble of up to 32 data permutations. These permutations are constructed using combinations of feature column rotations, class label rotations, and the application of a power transformation. To prevent redundant evaluations, the number of ensemble members is strictly bounded to the maximum number of unique pre-processing combinations, $2kj$, where $k$ is the number of features and $j$ is the number of classes. Unless otherwise noted, all reported evaluations utilize this ensembling approach, which yields marginal but consistent performance improvements.

\section{Model Configuration and Scaling}
\label{app:model-scaling}

The transformer backbone has four attention heads, pre-normalization, and no dropout. The feed-forward hidden dimension is double the embedding dimension, $d_{\text{hid}} = 2\times d_{\text{emb}}$.

\begin{figure}[htbp]
    \centering
    \begin{subfigure}[b]{0.48\textwidth}
        \centering
        \includegraphics[width=\textwidth]{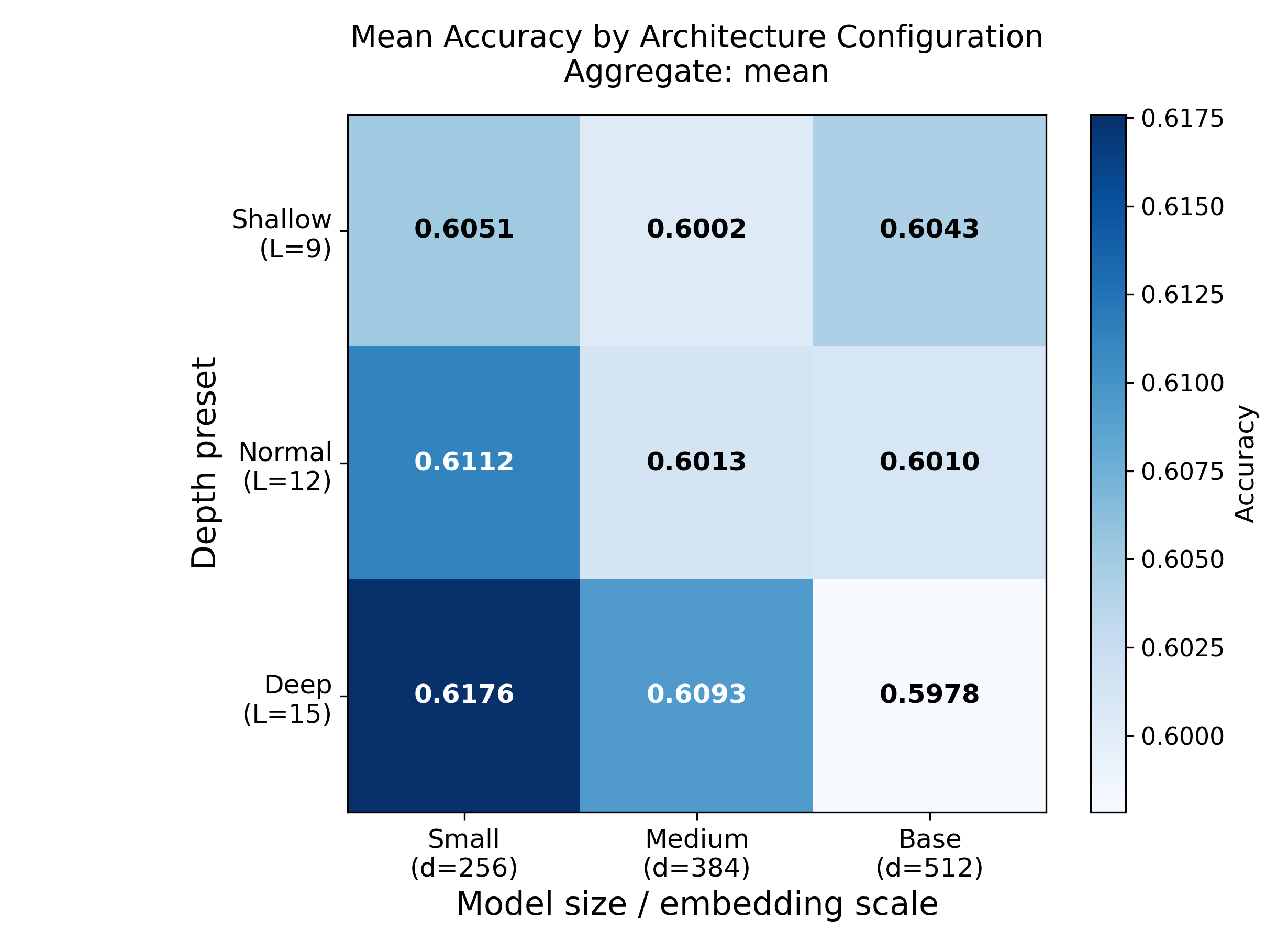}
        \caption{Accuracy}
        \label{fig:heatmap_acc}
    \end{subfigure}
    \hfill
    \begin{subfigure}[b]{0.48\textwidth}
        \centering
        \includegraphics[width=\textwidth]{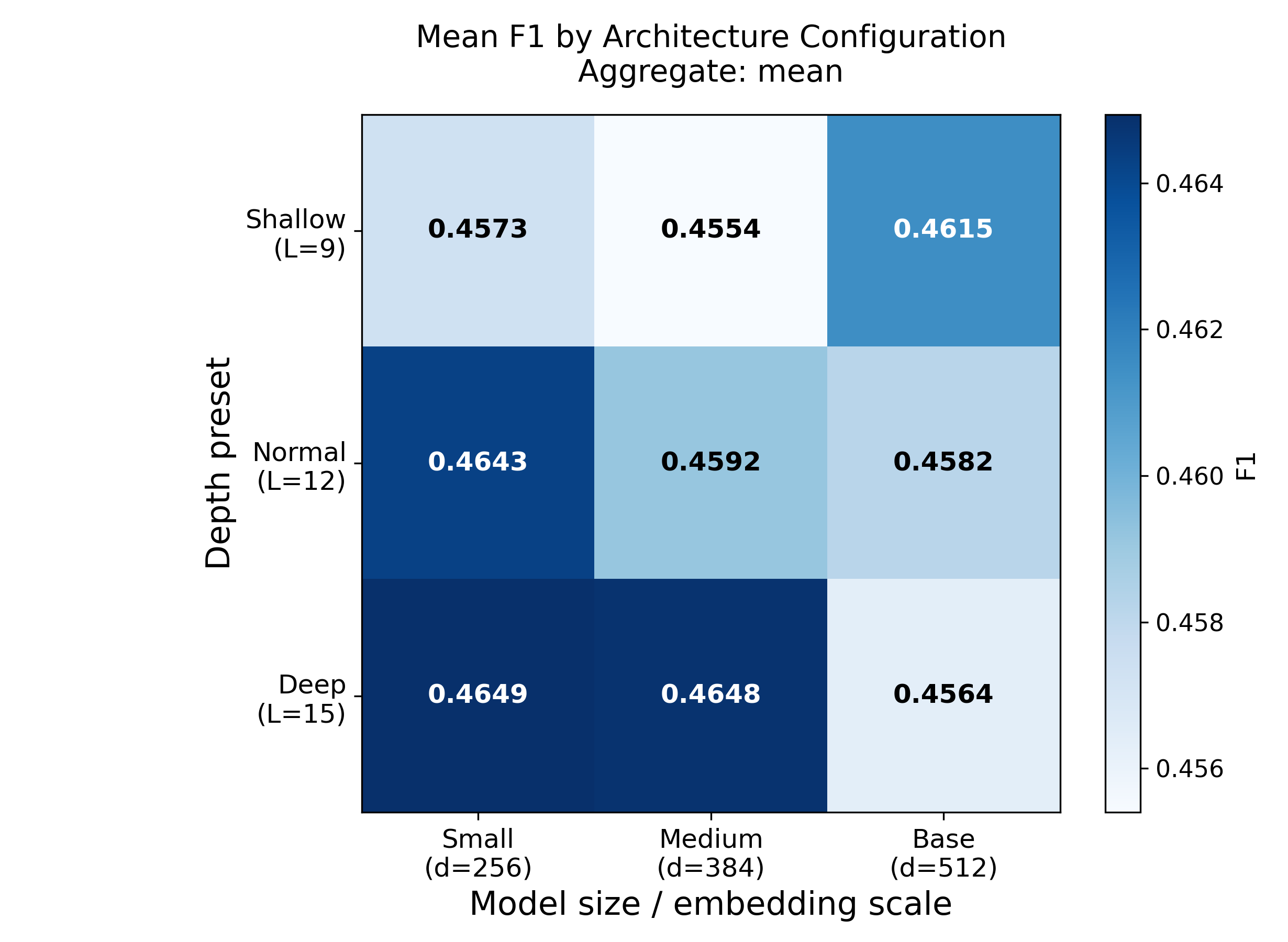}
        \caption{F1 Score}
        \label{fig:heatmap_f1}
    \end{subfigure}
    
    \vspace{1em}
    
    \begin{subfigure}[b]{0.48\textwidth}
        \centering
        \includegraphics[width=\textwidth]{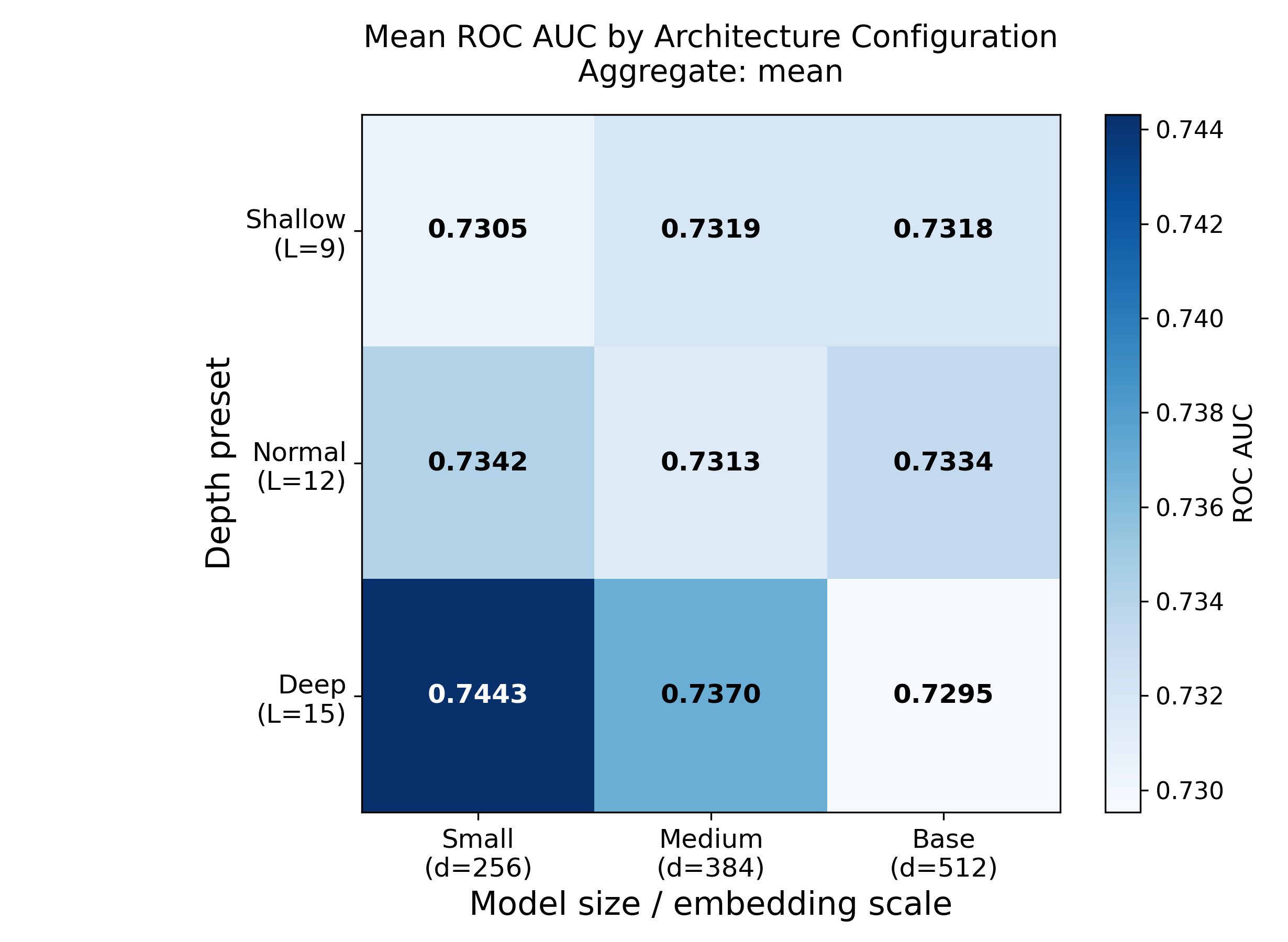}
        \caption{ROC AUC}
        \label{fig:heatmap_roc}
    \end{subfigure}
    \caption{Performance heatmaps for varying model configurations. The grid search evaluates 9 distinct architectures modifying the embedding dimension ($d_{\text{emb}}$) and number of layers ($L$). Across all 3 metrics, the 'Small Deep' configuration ($d_{\text{emb}}=256$, $L=15$) achieves the best performance.}
    \label{fig:scaling_heatmaps}
\end{figure}

We investigate a range of model configurations varying the embedding dimension $d_{\text{emb}}\in\{256, 384, 512\}$ and the number of layers $L\in\{9, 12, 15\}$. We conduct a grid-search across the nine configurations using the best prior from the prior configuration search (which used the median model architecture $d_{\text{emb}}=384$, $L=12$). For all model configurations, training spans 200 epochs so all architectures see the same number of training examples. Based on the loss curves, training has plateaued for all models by this point. We use the same scheduling, and the peak learning rate is adjusted logarithmically based on the number of parameters; details on both are described in \cref{app:training-details}. Full results are displayed below in \cref{fig:scaling_heatmaps}. 

\section{Training Details}
\label{app:training-details}

The training configuration is based on the hyperparameters used in TabPFN v1~\citep{hollmann_tabpfn_2023} with some adjustments. The synthetic datasets are sampled with 1,024 samples each and the evaluation position which separates the context and test points is sampled uniformly between position 0 and 1,000 for each dataset. Our model can handle a maximum of 100 features, 5 targets, and up to 10 classes per target.

The model is trained for 200 epochs of 128 batches each, with batch size 128 and gradient accumulation of 4. Thus the model sees 3,276,800 synthetic datasets throughout training. Training uses the Adam optimizer~\citep{Kingma2014AdamAM} with no weight decay and gradient clipping is applied with a maximum norm of 1.0 to ensure stability. The learning rate increases linearly over five epochs to a peak value of $9.379\times10^{-4}$ then decays following a cosine schedule. The peak learning rate was scaled according to the total parameter count of the network, utilizing the logarithmic curve-fit heuristic from \citet{hollmann_tabpfn_2023}, which approximates the optimal learning rates observed in large language models~\citep{brown_language_2020}.

Training uses mixed precision and flash attention to accelerate training and inference performance. All models were implemented in PyTorch. The models were trained using Distributed Data Parallel (DDP) across several NVIDIA RTX 5000 Ada (32 GB) GPUs. Hardware usage for hyperparameter sweeps and training runs varied between one and eight GPUs depending on the task. During training, the prior data is generated using a CPU with eight dedicated workers.

\section{Evaluation Dataset Details}
\label{app:dataset-details}

TabPFN-MT is evaluated using datasets from four distinct groups: Native Multitask, Single-Task (STL), Feature-Derived Multitask, and Subsampled Large-Scale Multitask. The coverage and sample size distributions for each category are detailed in \cref{tab:coverage}.

\begin{table}[ht]
\centering
\caption{Dataset coverage for model comparisons. The MTL evaluation spans 283 datasets with a diverse range of sample sizes, focusing on the low-data regime where $D$ indicates the number of datasets.}
\label{tab:coverage}
\setlength{\tabcolsep}{12pt}
\begin{tabular}{lrr}
\toprule
Source & $D$ & Samples ($\mu \pm \sigma$) \\
\midrule
STL & 61 & $689.1 \pm 794.4$ \\
\midrule
Native MTL & 11 & $1278.7 \pm 976.0$ \\
Derived MTL & 232 & $644.9 \pm 782.4$ \\
Large MTL & 40 & $2250.0 \pm 1772.3$ \\
MTL Avg. & 283 & $896.4 \pm 1134.6$ \\
\bottomrule
\end{tabular}
\end{table}

\subsection{Native Multitask Datasets}

Due to the lack of established multitask benchmarks in the small-to-medium data regime, we curated a collection of multi-target classification datasets from OpenML. We observed that many of these datasets exhibit extreme class imbalance, rendering them more akin to anomaly detection tasks than standard classification. To address this, we filtered the collection to exclude datasets with a maximum Imbalance Ratio (IR) greater than 16, defined as $\text{IR} = \frac{|\text{Majority Class}|}{|\text{Minority Class}|}$. Furthermore, to align with our experimental scope, we restricted the final selection to datasets containing a maximum of 5,000 samples, 100 features, and five targets.

\subsection{Standard Single-Task Datasets}

To evaluate performance on standard single-task classification problems ($T=1$), we utilized a subset of OpenML datasets drawn from the validation suite established by \citet{hollmann_tabpfn_2023}.

\subsection{Feature-Derived Multitask Datasets}
\label{app:derived-mtl}

To robustly evaluate model performance across a wider variety of domain structures and target correlations, we programmatically derived a suite of multi-target datasets from established single-target classification benchmarks. For each base dataset, we dropped the original target and repurposed a subset of the input features to act as a new joint target vector $\mathbf{y}$ of length $T$, where $T \in \{2, 3, 4, 5\}$.

To ensure diverse and representative task combinations, the transformation process alternates between selecting nominal and numeric features. 
\begin{itemize}
    \item \textbf{Nominal features} are extracted and treated directly as multiclass classification targets.
    \item \textbf{Numeric features} are cast into classification tasks via quantile discretization. To introduce varying levels of task complexity, the number of quantile bins systematically cycles through $b \in \{3, 4, 5, 6\}$. 
\end{itemize}

Crucially, once a feature is designated as a target, it is strictly removed from the input feature space $X$ to prevent trivial identity mapping. Datasets lacking sufficient features to support the extraction while maintaining at least one input feature were skipped. This methodology yields a challenging set of naturally occurring multitask environments where the generative relationships between the remaining features and the newly assigned targets are inherently complex and domain-grounded.

\subsection{Subsampled Large-Scale Datasets}

Additionally, we used two large scale multi-target datasets which have been common to evaluate deep tabular models: AliExpress~\citep{li_improving_2020} using the processed version from \citet{xi_modeling_2021} and ACS Income~\citep{ding_retiring_2021} using the 2 target setup from \citet{ma_modeling_2018}. However, because these full datasets contain millions of samples they are not suitable for our model as is, thus we subsample randomly from the datasets down to $\{500, 1000, 2500, 5000\}$ samples. We evaluate five random seeds for each size yielding 20 total subsampled datasets from each base dataset.

\section{Baseline Details}
\label{app:baseline-details}

To ensure a rigorous, fair, and reproducible evaluation, this section details the specific implementations and optimization protocols used for all baseline models compared against TabPFN-MT.

\subsection{Baseline Implementations}

We utilize PyTorch implementations from \citet{sinodinos_multitab_2026} for most deep models. For XGBoost, CatBoost, and TabPFN v2.6, we use their official libraries. For TabPFN v1, we use a public implementation and checkpoint, as the original weights are no longer accessible via the official TabPFN library.

\subsection{Baseline Hyperparameter Search}
\label{app:baseline-hp}

We conducted hyperparameter optimization for all baseline models and aggregated the results across 5 random seeds. Given the massive computational scale of evaluating across 150 datasets, all models were optimized using a constrained budget of 25 random search trials via Optuna, with each trial scored using 3-fold cross-validation (CV). To prevent the neural architectures from overfitting their training folds, our PyTorch estimators dynamically reserved a $10\%$ validation fraction from within each training fold to enable early stopping (patience = 10 epochs). Tree-based models, which are far less susceptible to epoch-based overfitting, were trained on the full CV training folds.

For XGBoost and CatBoost, we follow \citet{hollmann_tabpfn_2023} and utilize the hyperparameter search spaces established by \citet{shwartz-ziv_tabular_2022}. For Random Forest, we adopt a regularized search space, tuning from 50 to 300 estimators and restricting maximum depth from 3 to 15. Finally, for deep learning baselines, we specifically designed the search spaces to favor lower architectural capacities and heavier regularization (e.g., expanded ranges for dropout and weight decay) to prevent catastrophic overfitting in these low-data regimes. Full details on hyperparameter search spaces are included below in \cref{tab:baseline_hpo_spaces}.

\begin{table}[htbp]
\centering
\small
\setlength{\tabcolsep}{4pt}
\caption{Hyperparameter search spaces for tree-based and deep learning baselines. The bounds are adapted from the MultiTab framework. Static parameters used during deep baseline training include an early stopping patience of 10 epochs and a validation fraction of 10\%.}
\label{tab:baseline_hpo_spaces}
\begin{tabular}{lll}
\toprule
Hyperparameter & Distribution & Range / Choices \\
\midrule
\multicolumn{3}{l}{\textit{Random Forest}} \\
Number of estimators & Uniform Int & $[50, 300]$ \\
Max depth & Uniform Int & $[3, 15]$ \\
Min samples split & Uniform Int & $[2, 10]$ \\
Min samples leaf & Uniform Int & $[1, 5]$ \\
Max features & Categorical & $\{\text{sqrt}, \text{log2}, \text{None}\}$ \\
\midrule
\multicolumn{3}{l}{\textit{XGBoost}} \\
Number of estimators & Uniform Int & $[100, 4000]$ \\
Learning rate & LogUniform & $[10^{-7}, 1]$ \\
Max depth & Uniform Int & $[1, 10]$ \\
Subsample & Uniform & $[0.2, 1.0]$ \\
Colsample by tree & Uniform & $[0.2, 1.0]$ \\
Colsample by level & Uniform & $[0.2, 1.0]$ \\
Min child weight & LogUniform & $[10^{-16}, 10^5]$ \\
Alpha (L1 reg.) & LogUniform & $[10^{-16}, 10^2]$ \\
Lambda (L2 reg.) & LogUniform & $[10^{-16}, 10^2]$ \\
Gamma & LogUniform & $[10^{-16}, 10^2]$ \\
\midrule
\multicolumn{3}{l}{\textit{CatBoost}} \\
Learning rate & LogUniform & $[10^{-5}, 1]$ \\
Random strength & Uniform & $[1, 20]$ \\
One-hot max size & Uniform Int & $[0, 25]$ \\
L2 leaf reg & LogUniform & $[1, 10]$ \\
Bagging temperature & Uniform & $[0, 1]$ \\
Leaf estimation iterations & Uniform Int & $[1, 20]$ \\
\midrule
\multicolumn{3}{l}{\textit{Deep Learning Baselines}} \\
\multicolumn{3}{l}{\textit{(MultiTab, MT-MLP, MMoE, PLE, STEM, TabTransformer, SAINT, FT-Transformer)}} \\
Maximum epochs & Uniform Int & $[40, \text{max\_epochs}]$ \\
Batch size & Categorical & $\{128, 256, 512\}$ \\
Learning rate & LogUniform & $[10^{-4}, 5 \times 10^{-3}]$ \\
Weight decay & LogUniform & $[10^{-7}, 10^{-3}]$ \\
Dropout & Uniform & $[0.0, 0.4]$ \\
Embedding dimension & Categorical & $\{16, 32, 64\}$ \\
Hidden dimension & Categorical & $\{64, 128, 256\}$ \\
Number of blocks (layers) & Uniform Int & $[2, 4]$ \\
Number of attention heads & Categorical & $\{2, 4, 8\}$ \\
Feed-forward hidden dim & Categorical & $\{64, 128, 256\}$ \\
\bottomrule
\end{tabular}
\end{table}

\section{Extended Evaluation Details}
\label{app:extended-experiments}

\subsection{Multiclass Metric Aggregation}

The metric calculation in our experiments adapts dynamically depending on the cardinality of each target. For binary classification tasks, we report the standard binary F1 score and binary ROC AUC. For multiclass targets, we apply specific handling to ensure fair evaluation across varying and imbalanced class distributions: F1 scores are reported using a macro-average to equally weight all classes, and ROC AUC is calculated using the One-vs-One (OVO) macro-average approach, which computes the average AUC across all possible pairwise class combinations~\citep{hand_simple_2001}. 

Because our architecture jointly predicts varying numbers of tasks, dataset-level performance is evaluated by first computing these metrics independently for each of the $T$ targets. The overall performance for a given dataset is then reported as the unweighted mean of these individual target metrics. In instances where a metric is mathematically undefined for a specific target, it is excluded from the dataset's aggregated mean.

\subsection{Cross-Validation and Robustness} 

To ensure robustness against variances in data splitting and optimization, all baseline and proposed models undergo 5-fold cross-validation during evaluation. This entire training and evaluation pipeline is repeated across three independent random seeds. The final reported metrics for each dataset represent the mean performance across these repeated trials.

\subsection{Statistical Significance Testing}

To assess the statistical significance of our comparative results across a large corpus of datasets, we follow the non-parametric statistical pipeline recommended by \citet{demsar_statistical_2006}. We first apply the Friedman test to reject the null hypothesis that all models perform equivalently. Subsequently, we use the Nemenyi post-hoc test to perform pairwise comparisons between models. The results of these tests are visualized using Critical Difference (CD) diagrams, where models connected by a thick horizontal bar belong to the same top-performing "clique" and are not statistically significantly different from one another at a significance level of $\alpha = 0.05$.

\section{Extended Results}
\label{app:extended-results}

This section provides the complete, granular breakdown of the experimental results introduced in \cref{sec:experiment}. While the main text reports aggregated performance, \cref{tab:roc_auc}, \cref{tab:accuracy}, and \cref{tab:f1} stratify the results across our distinct data regimes described above in \cref{app:dataset-details}: standard Single-Task Learning (STL), Native MTL, Feature-Derived MTL, and Subsampled Large MTL.

For each metric, we report the average score alongside the average rank in the format of \textit{value (rank)}. Stratifying the models at this level reveals how performance shifts depending on the dataset's origin, scale, and inherent task correlation. For example, while deep multitask baselines show improved competitiveness on the Subsampled Large MTL datasets, they generally struggle in the low-data regimes. 

Conversely, TabPFN-MT maintains robust performance across all multitask categories. Notably, its strong performance on the Feature-Derived MTL datasets—where targets share an explicit, inherent causal relationship with the inputs—validates our synthetic data generation strategy, demonstrating that the model successfully captures structural dependencies similar to those generated by our single Structural Causal Model (SCM) prior.

\begin{table}[ht]
\small
\centering
\caption{Average ROC-AUC across dataset sources. Each cell reports average metric value and average rank as value (rank). Within each source column, \textbf{bold} marks the best value and \underline{underlined} text marks the second-best value (ties share the same formatting). For ranks, lower is better.}
\label{tab:roc_auc}
\setlength{\tabcolsep}{4pt}
\begin{tabular}{lrrrrr}
\toprule
Model & STL & Native MTL & Derived MTL & Large MTL & MTL Avg. \\
\cmidrule(lr){2-2} \cmidrule(lr){3-6}
Ours & 0.826 (5.03) & \textbf{0.788} (3.86) & \underline{0.717} (4.09) & \textbf{0.768} (2.65) & \underline{0.727} (3.87) \\
\midrule
\multicolumn{6}{l}{\textit{Multi-Task Baselines}} \\
MT-MLP & 0.790 (8.52) & 0.755 (8.45) & 0.677 (7.45) & 0.700 (6.28) & 0.683 (7.33) \\
MMoE & 0.769 (9.41) & 0.754 (7.91) & 0.667 (8.36) & 0.697 (7.40) & 0.675 (8.21) \\
PLE & 0.751 (10.85) & 0.727 (10.91) & 0.645 (10.14) & 0.672 (9.50) & 0.652 (10.08) \\
STEM & 0.737 (11.33) & 0.736 (10.18) & 0.643 (10.29) & 0.678 (9.28) & 0.652 (10.14) \\
MTab & 0.806 (8.04) & 0.752 (9.77) & 0.654 (9.51) & 0.705 (8.43) & 0.665 (9.36) \\
\midrule
\multicolumn{6}{l}{\textit{Single-Task Baselines}} \\
ST-MLP & 0.786 (8.67) & 0.763 (8.23) & 0.677 (8.20) & 0.652 (11.50) & 0.677 (8.67) \\
RF & \underline{0.852} (5.05) & 0.773 (5.41) & 0.706 (6.00) & 0.706 (5.65) & 0.708 (5.93) \\
XGB & 0.768 (11.05) & 0.710 (13.45) & 0.659 (10.49) & 0.668 (10.82) & 0.662 (10.65) \\
CatB & 0.848 (4.31) & 0.774 (4.77) & 0.704 (6.24) & 0.723 (6.05) & 0.709 (6.15) \\
Tab-T & 0.761 (11.13) & 0.748 (8.45) & 0.644 (10.50) & 0.653 (11.25) & 0.649 (10.52) \\
FT-T & 0.806 (8.04) & 0.746 (10.23) & 0.671 (8.95) & 0.688 (9.03) & 0.676 (9.01) \\
SAINT & 0.732 (12.20) & 0.698 (13.18) & 0.633 (11.97) & 0.612 (13.70) & 0.632 (12.27) \\
TabPFN-v1 & 0.834 (4.15) & 0.783 (2.68) & 0.715 (4.50) & 0.733 (4.85) & 0.720 (4.48) \\
TabPFN-v2.6 & \textbf{0.877} (2.21) & \underline{0.786} (2.50) & \textbf{0.731} (3.31) & \underline{0.733} (3.62) & \textbf{0.734} (3.33) \\
\bottomrule
\end{tabular}
\end{table}

\begin{table}[ht]
\small
\centering
\caption{Average Accuracy across dataset sources. Each cell reports average metric value and average rank as value (rank). Within each source column, \textbf{bold} marks the best value and \underline{underlined} text marks the second-best value (ties share the same formatting). For ranks, lower is better.}
\label{tab:accuracy}
\setlength{\tabcolsep}{4pt}
\begin{tabular}{lrrrrr}
\toprule
Model & STL & Native MTL & Derived MTL & Large MTL & MTL Avg. \\
\cmidrule(lr){2-2} \cmidrule(lr){3-6}
Ours & 0.805 (4.29) & 0.645 (6.09) & \textbf{0.526} (4.73) & 0.864 (5.49) & \textbf{0.578} (4.89) \\
\midrule
\multicolumn{6}{l}{\textit{Multi-Task Baselines}} \\
MT-MLP & 0.741 (9.56) & 0.603 (8.55) & 0.455 (7.75) & 0.864 (5.29) & 0.518 (7.43) \\
MMoE & 0.727 (10.08) & 0.599 (9.50) & 0.439 (8.96) & 0.861 (6.44) & 0.505 (8.62) \\
PLE & 0.720 (10.63) & 0.576 (11.73) & 0.417 (10.55) & 0.859 (6.94) & 0.486 (10.08) \\
STEM & 0.708 (11.39) & 0.574 (10.64) & 0.417 (10.36) & 0.859 (7.83) & 0.486 (10.01) \\
MTab & 0.774 (7.82) & 0.625 (8.32) & 0.451 (8.67) & 0.853 (6.91) & 0.515 (8.41) \\
\midrule
\multicolumn{6}{l}{\textit{Single-Task Baselines}} \\
ST-MLP & 0.765 (8.25) & 0.633 (8.05) & 0.476 (6.79) & 0.856 (9.01) & 0.536 (7.15) \\
RF & \underline{0.816} (5.75) & 0.645 (4.91) & \underline{0.509} (4.73) & \underline{0.867} (4.84) & \underline{0.565} (4.75) \\
XGB & 0.694 (10.98) & 0.546 (12.91) & 0.416 (11.55) & 0.684 (13.91) & 0.459 (11.93) \\
CatB & 0.810 (5.70) & 0.644 (5.59) & 0.507 (5.03) & 0.730 (10.34) & 0.544 (5.80) \\
Tab-T & 0.738 (9.93) & 0.609 (7.36) & 0.436 (9.75) & 0.856 (8.91) & 0.502 (9.54) \\
FT-T & 0.774 (7.82) & 0.622 (9.36) & 0.452 (8.62) & 0.856 (9.29) & 0.516 (8.75) \\
SAINT & 0.727 (10.22) & 0.592 (11.82) & 0.419 (11.26) & 0.828 (10.68) & 0.483 (11.20) \\
TabPFN-v1 & 0.805 (4.65) & \underline{0.652} (3.14) & 0.486 (6.20) & 0.730 (9.82) & 0.527 (6.60) \\
TabPFN-v2.6 & \textbf{0.849} (2.94) & \textbf{0.659} (2.05) & 0.503 (5.05) & \textbf{0.874} (4.31) & 0.562 (4.83) \\
\bottomrule
\end{tabular}
\end{table}

\begin{table}[ht]
\small
\centering
\caption{Average F1 across dataset sources. Each cell reports average metric value and average rank as value (rank). Within each source column, \textbf{bold} marks the best value and \underline{underlined} text marks the second-best value (ties share the same formatting). For ranks, lower is better.}
\label{tab:f1}
\setlength{\tabcolsep}{4pt}
\begin{tabular}{lrrrrr}
\toprule
Model & STL & Native MTL & Derived MTL & Large MTL & MTL Avg. \\
\cmidrule(lr){2-2} \cmidrule(lr){3-6}
Ours & 0.674 (4.70) & \underline{0.521} (6.41) & 0.434 (4.50) & 0.319 (7.47) & \underline{0.421} (5.00) \\
\midrule
\multicolumn{6}{l}{\textit{Multi-Task Baselines}} \\
MT-MLP & 0.545 (9.37) & 0.419 (7.73) & 0.344 (8.67) & \underline{0.323} (5.94) & 0.344 (8.25) \\
MMoE & 0.519 (9.84) & 0.406 (9.27) & 0.320 (9.94) & 0.322 (6.09) & 0.324 (9.37) \\
PLE & 0.481 (10.95) & 0.361 (10.45) & 0.286 (11.62) & 0.315 (7.58) & 0.293 (11.00) \\
STEM & 0.478 (11.02) & 0.354 (10.00) & 0.286 (11.54) & 0.313 (6.92) & 0.292 (10.83) \\
MTab & 0.630 (7.88) & 0.462 (9.23) & 0.330 (10.03) & 0.302 (8.79) & 0.331 (9.82) \\
\midrule
\multicolumn{6}{l}{\textit{Single-Task Baselines}} \\
ST-MLP & 0.658 (6.30) & 0.520 (4.14) & 0.406 (5.46) & 0.317 (7.56) & 0.398 (5.71) \\
RF & \underline{0.691} (5.27) & 0.514 (4.36) & \textbf{0.447} (3.11) & 0.322 (6.19) & \textbf{0.432} (3.59) \\
XGB & 0.396 (12.36) & 0.273 (13.27) & 0.292 (10.77) & 0.106 (12.64) & 0.265 (11.13) \\
CatB & 0.679 (5.27) & 0.509 (5.05) & \underline{0.445} (3.34) & 0.237 (10.09) & 0.418 (4.36) \\
Tab-T & 0.564 (10.25) & 0.457 (8.36) & 0.333 (8.97) & 0.312 (9.24) & 0.335 (8.98) \\
FT-T & 0.630 (7.88) & 0.462 (9.45) & 0.332 (9.44) & 0.317 (7.78) & 0.335 (9.21) \\
SAINT & 0.518 (11.16) & 0.351 (12.36) & 0.278 (12.15) & 0.260 (10.70) & 0.278 (11.95) \\
TabPFN-v1 & 0.665 (4.66) & 0.510 (6.41) & 0.403 (5.73) & 0.242 (8.61) & 0.384 (6.16) \\
TabPFN-v2.6 & \textbf{0.730} (3.10) & \textbf{0.536} (3.50) & 0.425 (4.72) & \textbf{0.337} (4.41) & 0.417 (4.63) \\
\bottomrule
\end{tabular}
\end{table}

\begin{figure}[htbp]
    \centering
    \begin{subfigure}{\linewidth}
        \centering
        \includegraphics[width=0.9\linewidth]{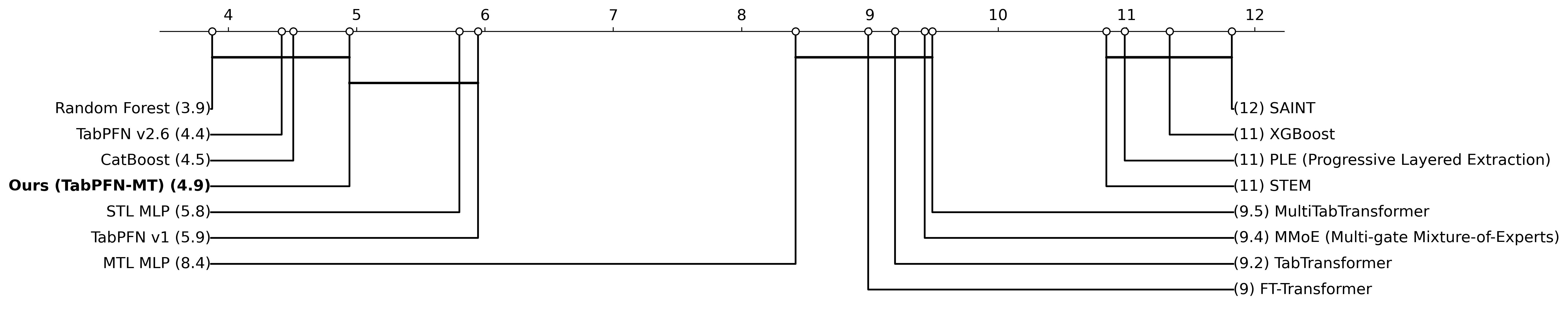}
        \caption{\textbf{F1 Score:} TabPFN-MT belongs to the top statistical clique and is the highest-ranked multitask model.}
        \label{fig:cd-f1}
    \end{subfigure}
    
    \vspace{1.5em}
    
    \begin{subfigure}{\linewidth}
        \centering
        \includegraphics[width=0.9\linewidth]{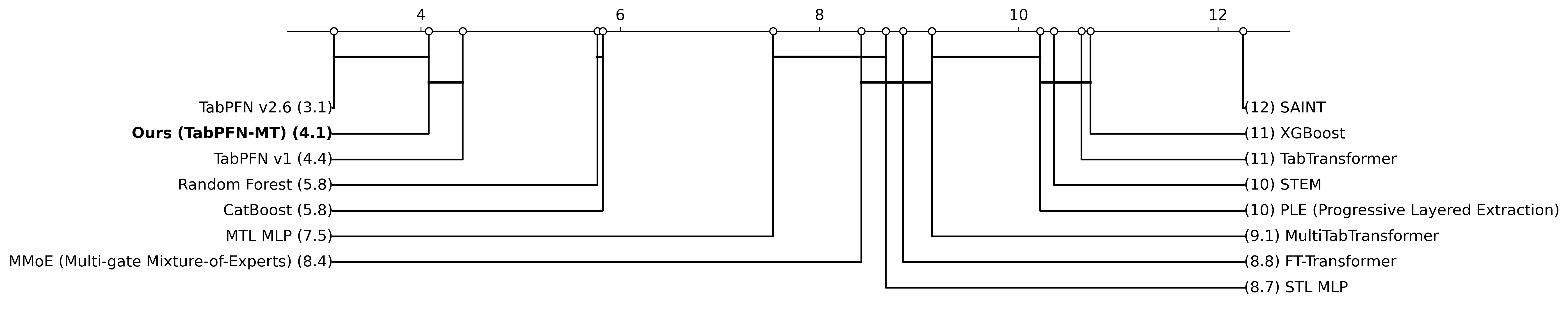}
        \caption{\textbf{ROC AUC:} TabPFN-MT ranks second overall and shares the top-performing clique with TabPFN v2.6.}
        \label{fig:cd-roc_auc}
    \end{subfigure}
    
    \caption{\textbf{Critical difference diagrams for auxiliary metrics.} Consistent with accuracy results, TabPFN-MT significantly outperforms all other evaluated multitask baselines across both F1 and ROC AUC. Notably, for both metrics, our model is not significantly different from the top-ranked baselines, placing it in the highest statistical tier.}
    \label{fig:cd-all}
\end{figure}

\subsection{Predictive Performance vs. Target Count}
\label{app:pred-vs-capacity}

\begin{figure}
    \centering
    \includegraphics[width=0.95\linewidth]{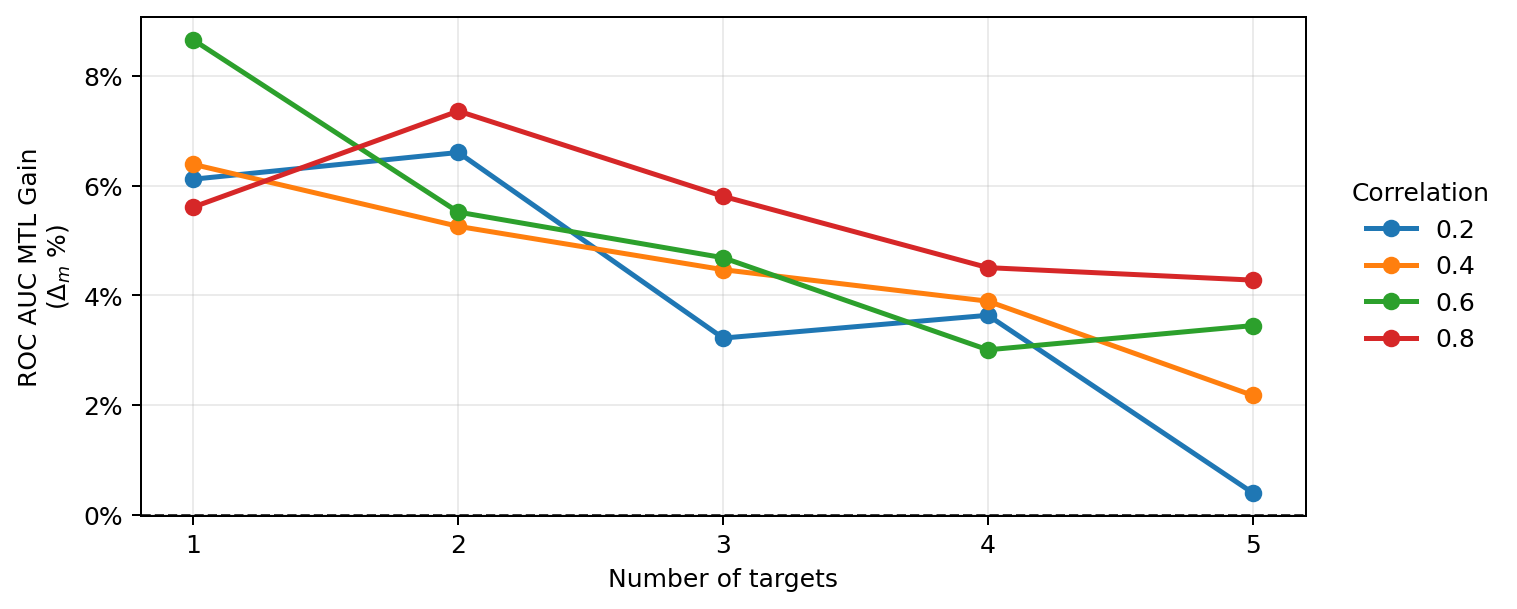}
    \caption{
    Effect of task scaling and correlation on multitask gain. We report the relative ROC AUC improvement ($\Delta_m \%$) of TabPFN-MT compared to a single-task baseline. As the model divides its fixed capacity across more targets, the multitask performance advantage gradually decreases. This drop-off is most severe for datasets with low inter-task correlation ($\text{correlation}=0.2$), indicating an information bottleneck when modelling weakly related targets.}
    \label{fig:mtl-gain-by-num-tasks}
\end{figure}

To better understand the capacity limitations of joint inference, we ablate the effect of increasing the number of targets on predictive performance by sweeping the task correlation from 0.2 to 0.8. In \cref{fig:mtl-gain-by-num-tasks} we see that the model's performance relative to a single-task baseline tends to gradually decrease as the number of tasks increases. This makes sense as TabPFN-MT must divide its capacity across modelling each target whereas the baseline uses increasing levels of compute as the number of tasks increases. We observe that the drop-off in performance is more severe in cases with lower task correlation. This does indicate that there is an information bottleneck in the decoder during joint inference which could motivate different approaches to decoding in future work. 

\section{Extended Computational Efficiency Analysis}
\label{app:extended-efficiency}

This section expands upon the computational efficiency findings discussed in \cref{sec:comp-scaling}. To rigorously evaluate the cost of multi-target scaling across diverse model architectures, we utilize two complementary metrics: a hardware-agnostic FLOPs estimation for deep learning models (\cref{app:flops-methodology}), and an end-to-end wall-clock runtime analysis (\cref{app:runtime-efficiency}) designed to capture the practical optimization overhead of tree-based baselines like GBDTs.

\subsection{FLOPs Estimation Methodology}
\label{app:flops-methodology}

For a rigourous and hardware-agnostic comparison of computational cost, we measure the total Floating Point Operations (FLOPs) required for the training and inference of each architecture. We use Pytorch's native profiler (\verb|torch.profiler|) for all models.

Our estimation procedure is defined as follows:
\begin{enumerate}
    \item \textbf{Hyperparameter Optimization (HPO):} For all trainable baselines, we first execute an HPO search via Optuna to determine the optimal, dataset-specific architecture (e.g., layer depth, embedding dimensions) and maximum training epochs. Note that the FLOPs required to conduct this search are omitted from our final figures; our reported FLOPs represent only the cost of the single, final optimal fit, making our comparison deliberately conservative in favor of the baselines.
    \item \textbf{Representative Profiling:} We isolate a single, representative forward batch and a single complete training step (forward pass, loss calculation, backward pass, and optimizer step). We profile these steps using `torch.profiler`, recording the median FLOPs across multiple trials to ensure stability.
    \item \textbf{Extrapolation:} The profiled training step FLOPs are multiplied by the total number of batches per epoch and the true number of executed epochs (accounting for early stopping). Inference FLOPs are similarly extrapolated by scaling the forward-batch FLOPs by the total required inference batches. 
    \item \textbf{multitask Accounting:} For single-target baselines applied to multi-target datasets, independent models are instantiated, tuned, and evaluated for each of the $T$ targets. Their individual computational footprints are summed, explicitly demonstrating the $\mathcal{O}(T)$ scaling penalty. Native multitask models process all targets jointly, requiring only a single training and inference cycle ($\mathcal{O}(1)$).
\end{enumerate}

\subsection{Runtime Analysis}
\label{app:runtime-efficiency}

While FLOPs are a hardware-agnostic metric, they do not capture the performance of tree-based models like GBDTs, which rely on logical splits rather than matrix multiplications. To capture end-to-end pipeline efficiency, we conduct a supplementary wall-clock runtime analysis.

All runtime benchmarking experiments were executed on a single NVIDIA RTX 5000 GPU utilizing 16 CPU threads. To mitigate timing variance from hardware initialization and data loading overhead, each evaluation is repeated three times, and the median runtime is reported.

\begin{figure}[t]
    \centering
    \includegraphics[width=0.95\linewidth]{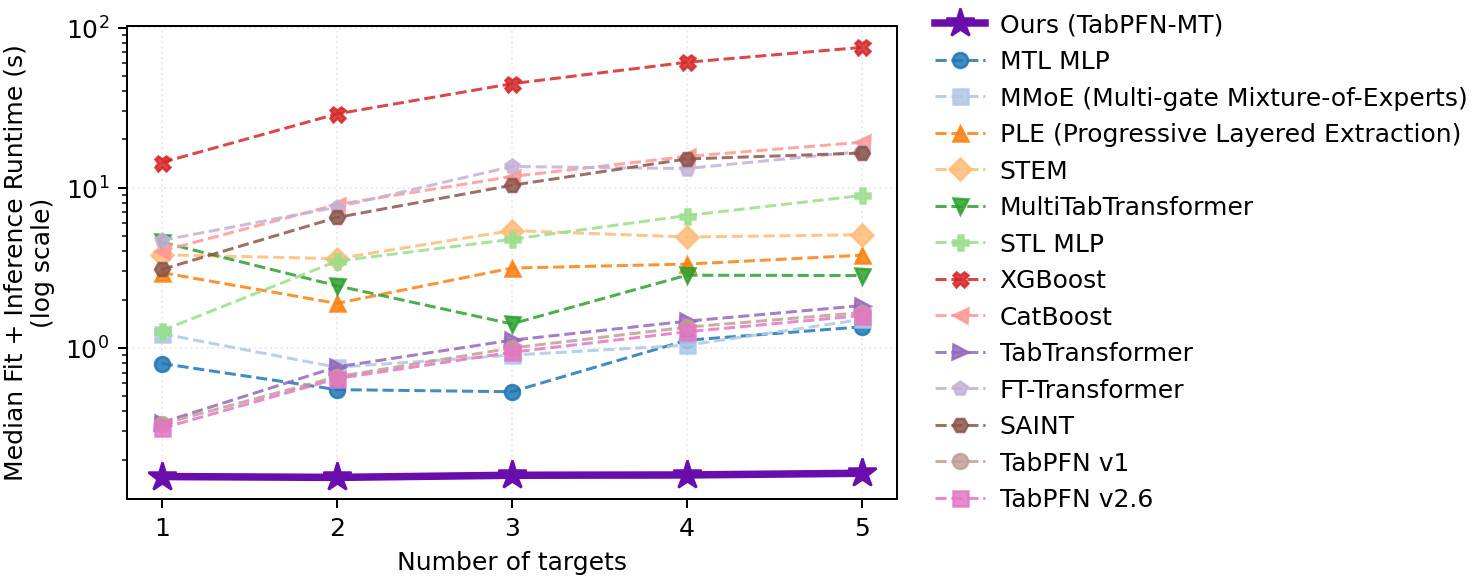}
    \caption{Cost of target scaling across architectures. Total runtime (fit and inference) is evaluated on synthetic datasets containing 1 to 5 targets. While single-target baselines exhibit linear scaling penalties ($\mathcal{O}(T)$), our proposed multitask architecture maintains near-constant execution time ($\mathcal{O}(1)$), significantly outperforming both single- and multi-target baselines.}
    \label{fig:comp-cost-scaling}
\end{figure}

As illustrated in \cref{fig:comp-cost-scaling}, single-target baselines incur a linear wall-clock penalty ($\mathcal{O}(T)$) as the target space grows. When scaling from 1 to 5 targets, single-task methods (including GBDTs) have a scaling factor around $5.0\times$, because they require independent HPO, fitting, and inference for every variable. multitask deep learning baselines process targets jointly, but still incur optimization overhead that scales with the complexity of the joint loss landscape and the necessary HPO phase. 

Because TabPFN-MT requires absolutely no dataset-specific gradient updates or HPO search, its wall-clock scaling factor remains near $1.0\times$ ($\mathcal{O}(1)$) regardless of target count. A detailed breakdown of these computational costs, including the exact wall-clock penalties introduced by the HPO phase across varying target counts, is provided in \cref{tab:timing_t1_t5_scaling}. 

\begin{table}[t]
\centering
\small
\caption{Median per-dataset timing comparison for T=1 vs T=5 with scaling.}
\label{tab:timing_t1_t5_scaling}
\setlength{\tabcolsep}{5pt}
\begin{tabular}{lrrrrrrrrr}
\toprule
 & \multicolumn{2}{c}{HPO (s)} & \multicolumn{2}{c}{Train (s)} & \multicolumn{2}{c}{Infer (s)} & \multicolumn{2}{c}{Total (s)} & Scaling \\
 \cmidrule(lr){2-3} \cmidrule(lr){4-5} \cmidrule(lr){6-7} \cmidrule(lr){8-9}
Model & T=1 & T=5 & T=1 & T=5 & T=1 & T=5 & T=1 & T=5 & Total x \\
\midrule
Ours & 0.00 & 0.00 & 0.00 & 0.00 & 0.01 & 0.02 & \textbf{0.01} & \textbf{0.02} & \textbf{1.44x} \\
\midrule
\multicolumn{10}{l}{\textit{Multi-Task Baselines}} \\
MT-MLP & 11.58 & 19.41 & 0.79 & 1.35 & 0.00 & 0.00 & 12.49 & 20.91 & 1.67x \\
MMoE & 12.11 & 18.38 & 1.22 & 1.50 & 0.00 & 0.00 & 13.33 & 19.88 & \underline{1.49x} \\
PLE & 13.40 & 26.82 & 2.94 & 3.78 & 0.01 & 0.01 & 16.34 & 30.24 & 1.85x \\
STEM & 16.59 & 46.05 & 3.80 & 5.07 & 0.01 & 0.01 & 20.54 & 50.48 & 2.46x \\
MTab & 20.29 & 44.92 & 4.56 & 2.82 & 0.01 & 0.00 & 24.86 & 46.45 & 1.87x \\
\midrule
\multicolumn{10}{l}{\textit{Single-Task Baselines}} \\
ST-MLP & 10.66 & 54.81 & 1.28 & 8.92 & 0.00 & 0.01 & 11.94 & 63.74 & 5.34x \\
XGB & 184.23 & 905.62 & 14.29 & 74.91 & 0.07 & 0.28 & 198.57 & 980.61 & 4.94x \\
CatB & 149.80 & 707.04 & 4.01 & 19.18 & 0.01 & 0.03 & 153.80 & 726.21 & 4.72x \\
Tab-T & 3.60 & 20.07 & 0.34 & 1.83 & 0.00 & 0.00 & 3.95 & 21.89 & 5.54x \\
FT-T & 19.51 & 125.78 & 4.68 & 16.79 & 0.01 & 0.04 & 24.23 & 144.29 & 5.95x \\
SAINT & 61.65 & 309.24 & 3.07 & 16.38 & 0.01 & 0.06 & 64.71 & 325.27 & 5.03x \\
TabPFN-v1 & 0.00 & 0.00 & 0.00 & 0.00 & 0.33 & 1.66 & 0.33 & 1.66 & 4.96x \\
TabPFN-v2.6 & 0.00 & 0.00 & 0.06 & 0.31 & 0.25 & 1.27 & \underline{0.31} & \underline{1.59} & 5.06x \\
\bottomrule
\end{tabular}
\end{table}

%%%%%%%%%%%%%%%%%%%%%%%%%%%%%%%%%%%%%%%%%%%%%%%%%%%%%%%%%%%%

% \newpage
% \input{checklist.tex}

\end{document}